\newif\nohyperref
\documentclass[]{article}

\newif\ifispublish
\ispublishtrue

\usepackage{microtype}
\usepackage{graphicx}
\usepackage{booktabs} % for professional tables

\usepackage{hyperref}

\usepackage[accepted]{mlsys2025}
\makeatletter
\providecommand{\mlsys@appearing}{}
\makeatother

\usepackage[nolist,nohyperlinks]{acronym}
\usepackage[utf8]{inputenc} % allow utf-8 input
\usepackage[T1]{fontenc}    % use 8-bit T1 fonts
\usepackage{hyperref}       % hyperlinks
\usepackage{xurl}           % allow long bibliography URLs to break in columns
\usepackage{booktabs}       % professional-quality tables
\usepackage{array}          % column specifiers (>{...})
\usepackage{amsfonts}       % blackboard math symbols
\usepackage{nicefrac}       % compact symbols for 1/2, etc.
\usepackage{microtype}      % microtypography
\usepackage{xcolor}         % colors

\usepackage{cuted}
\usepackage{capt-of}

\usepackage{xspace}
\usepackage{listings}
\usepackage{xcolor}

\usepackage{amsmath,amssymb}
\usepackage{algorithm}
\usepackage{float}
\usepackage{setspace}

\lstdefinestyle{mintedlike}{
  language=Python,
  basicstyle=\ttfamily\footnotesize,
  stringstyle=\color{red},
  keywordstyle=\color{black},
  commentstyle=\color{black},
  identifierstyle=\color{black},
  showstringspaces=false,
  breaklines=false,
  frame=none,
  xleftmargin=0pt,
  framexleftmargin=0pt,
  aboveskip=0pt,
  belowskip=0pt,
}
\usepackage[most]{tcolorbox}
\newtcolorbox{codebox}[1][]{
  colback=black!3,
  colframe=black!3,
  boxrule=0pt,
  arc=0pt,
  left=2pt, right=2pt, top=2pt, bottom=2pt,
  height=2.8cm,          % fixed height — same for both
  valign=top,
  #1
}

\usepackage{subcaption}
\usepackage{multirow}

\ifispublish
    \usepackage[disable]{todonotes}
\else
    \usepackage{fancyhdr}
    \usepackage[colorinlistoftodos]{todonotes}
\fi

\usepackage[normalem]{ulem}
\useunder{\uline}{\ul}{}

\usepackage[capitalise]{cleveref}
\crefname{line}{line}{lines}
\crefalias{lstnumber}{line}
\makeatletter
\lst@AddToHook{Init}{\def\@currentcounter{lstnumber}}
\AddToHook{cmd/appendix/before}{\def\cref@section@alias{appendix}}
\makeatother

\usepackage{comment}

\usepackage{afterpage}

\newcommand{\para}[1]{\vspace{-3pt}\noindent\textbf{\textit{#1}.}}
\renewcommand{\cite}[1]{\citep{#1}}

\newcommand{\eg}{e.g.,\xspace}

\newcommand{\scode}[1]{{\small \texttt{#1}}}

\newcommand{\ignore}[1]{}

\ifispublish
     \newcommand{\TODO}[1]{}
     \newcommand{\assign}[2]{}

     \newcommand{\red}[1]{{#1}}
     
 \else
     \newcommand{\TODO}[1]{{\textcolor{red}{[\textbf{TODO:} #1]}}}
     \newcommand{\red}[1]{\textcolor{red}{#1}}
     
     \newcommand{\assign}[2]{{\colorbox{cyan}{[for \textbf{#1:} #2]}}}
 \fi

\newcommand\tabletextsize{\footnotesize}

\def\addauthnote#1#2{%
	\expandafter\def\csname#1\endcsname##1{%
	\todo[size=\footnotesize,color=#2,inline]
			{\textbf{\underline{\texttt{#1}}:} ##1}\xspace}
   \expandafter\def\csname#1li\endcsname##1{%
    \todo[size=\footnotesize,color=#2,inline,inlinewidth=5cm, noinlinepar]
			{\textbf{\underline{\texttt{#1}}:} ##1}\xspace}
    \expandafter\def\csname#1Res\endcsname##1{%
	\todo[size=\footnotesize,color=gray,inline]
			{\textbf{\underline{[Resolved] \texttt{#1}}:} ##1}\xspace}
       \expandafter\def\csname#1liRes\endcsname##1{%
    \todo[size=\footnotesize,color=gray,inline,inlinewidth=5cm, noinlinepar]
			{\textbf{\underline{\texttt{#1}}:} ##1}\xspace}
}

\def\shac#1{
	\expandafter\def\csname#1\endcsname{\ac{#1}\xspace}
    \expandafter\def\csname#1c\endcsname{\Ac{#1}\xspace}{}
    \expandafter\def\csname#1s\endcsname{\acp{#1}\xspace}{}
}

\newcommand{\sys}{{\textsc{OrigaMI} }\xspace}  % Michael's system for MI! (Working title) or is it a framework?
\newcommand{\NNSIGHT}{NNsight}
\newcommand{\TLENS}{TransformerLens}
\def\ptitle{OrigaMI: High-Performance Serving-Time Platform for Mechanistic Interpretability Apps}
\addauthnote{yaniv}{green!20}
\addauthnote{michael}{violet!25}
\addauthnote{mark}{cyan!25}

\addauthnote{TODO}{red!45}

\newcommand{\VLLMNS}{\texttt{vLLM}}
\newcommand{\VLLM}{\VLLMNS\xspace}
\newcommand{\ES}{\texttt{EasySteer}\xspace}

\newcommand{\MIMethod}{MI method\xspace}
\newcommand{\MIMethods}{MI methods\xspace}
\newcommand{\MIapps}{MI apps\xspace}
\newcommand{\MIapp}{MI app\xspace}
\newcommand{\FI}{FlashInspect\xspace}

\newcommand{\uncond}{-} %Uncond.

\newcommand{\TotalAppCount}{{20}\xspace}
\newcommand{\RepAppCount}{{10}\xspace}
\newcommand{\tbd}[1]{\textcolor{red}{[#1]}}

\ifispublish
    
\else
\fi

\mlsystitlerunning{OrigaMI: An LLM Serving System with First-Class Interpretability Apps}

\begin{document}

% Acronym definitions. With the `nolist' option this emits no text, so it is
% safe to place before the \twocolumn[...] title block (the abstract uses it).
\begin{acronym}

\acro{CFG}{control flow graph}
\acro{DFA}{deterministic finite automaton}
\acro{DSL}{domain specific language}
\acro{FSM}{finite state machine}
\acro{PTA}{prefix tree acceptor}
\acro{LLM}{large language model}
\acro{MLP}{multilayer perceptron}
\acro{MI}{mechanistic interpretability}
\acro{SAE}{sparse autoencoder}
\acro{MoE}{mixture of experts}
\acro{TTFT}{time to first token}
\acro{TTT}{total token throughput}
\acro{ITL}{inter-token latency}
\acro{FA}{FlashAttention}

\end{acronym}

\acrodefplural{DFA}[DFAs]{deterministic finite automata}
\acrodefplural{FSM}[FSMs]{finite state machines}
\acrodefplural{LLM}[LLMs]{large language models}
\acrodefplural{MLP}[MLPs]{multilayer perceptrons}
\acrodefplural{SAE}[SAEs]{sparse autoencoders}
\acrodefplural{MoE}[MoEs]{mixture of experts}

\shac{FA}
\shac{LLM}
\shac{TTFT}
\shac{TTT}
\shac{ITL}
\shac{MI}
\shac{DSL}

\twocolumn[
\mlsystitle{\ptitle}

% unless you've provided the [accepted] option to the mlsys2025 package.

\begin{mlsysauthorlist}
\mlsysauthor{Michael Blum}{tech}
\mlsysauthor{Mark Silberstein}{tech}
\mlsysauthor{Yaniv David}{tech}
\end{mlsysauthorlist}

\mlsysaffiliation{tech}{Technion -- Israel Institute of Technology, Haifa, Israel}

% Hidden during blind review; required for the camera-ready version.
% Change the name/address if someone else should be the contact author.
\mlsyscorrespondingauthor{Michael Blum}{michael.blum@campus.technion.ac.il}

% Keywords populate the "keywords" PDF metadata but are not shown in the
% document. See \pkeywords in Common/specificmacros.tex (currently empty).
\mlsyskeywords{}

\vskip 0.3in

\begin{abstract}
Mechanistic interpretability (MI) enables monitoring and steering LLMs by
operating on their internal state during inference. MI applications
remain difficult to develop, expensive to execute, and challenging to manage
at runtime.

We present \textbf{\sys}, a system that integrates MI applications as
first-class components of LLM serving, enabling simplified development,
efficient execution, and flexible deployment. Guided by a comprehensive
analysis of modern MI applications, \sys streamlines the \emph{\MIapps}
development via a simple Python-based Domain Specific Language (DSL) for
accessing intra- and inter-layer runtime model state at well-defined trigger
points during inference. \sys enables efficient access to ephemeral attention
scores via a novel \FI kernel that retains the performance benefits of
FlashAttention.
\sys's compiler integrates multiple \MIapps into the performance-optimized
inference execution flow while ensuring conflict-free coexistence. \sys's
runtime supports dynamic, per-request activation and deactivation of \MIapps
without service interruption or performance degradation when inactive,
leveraging \sys' dynamic memory management and CUDA graph-compatible
execution.

We integrate \sys with \VLLM and evaluate ten \MIapps, including sparse
autoencoders, steering, and jailbreak detection, across four
state-of-the-art LLMs. \sys incurs negligible execution overhead, up to two
orders of magnitude lower than existing frameworks, while enabling per-request
activation and deactivation up to 1,000 times faster than alternatives.
These results demonstrate that \sys makes MI applications practical for
deployment in production LLM serving systems.
\end{abstract}

]


\printAffiliationsAndNotice{}  % leave blank if no need to mention equal contribution

\section{Introduction}
\label{SEC:introduction}
Mechanistic Interpretability (MI) has emerged as a valuable approach for analyzing, monitoring, and modifying LLM behavior at inference time. Recent research demonstrates its effectiveness across a wide range of applications, including hallucination detection~\cite{orgad2025llmsknowshowintrinsic}, conditional refusal~\cite{programmingrefusalconditionalactivation}, safety steering~\cite{stickland2024steering}, and reasoning calibration~\cite{chen2025sealsteerablereasoningcalibration}. Moreover, \MI has been used to protect deployed language models, for example, by detecting personal information leakage using sparse autoencoders~\cite{goodfireRakutenSAE2025} and deploying misuse-detection probes in user-facing Gemini~\cite{kramarGeminiProbes2026}. These efforts demonstrate that inspecting and intervening in internal model state can not only explain model behavior but also inform operational decisions.

Operational experience suggests that deploying and dynamically toggling multiple \MI functions during model execution is increasingly important. For example, Google's Gemini study shows that input distribution shifts requires using different probes to maintain effectiveness~\cite{kramarGeminiProbes2026}. Similarly, Anthropic has evaluated classifier cascades that invoke progressively sophisticated \MI probes on production traffic, incurring additional overhead only for suspicious requests~\cite{cunninghamConstitutionalClassifiers2026}. Its pre-release model audits also employ different combinations of \MI functions~\cite{anthropic2025sonnetcard,anthropic2026mythoscard}.

Motivated by our analysis of \TotalAppCount prior \MIMethods
(\cref{SEC:abstraction}) and emerging production use cases, we posit that
\MI is becoming a new class of \emph{post-deployment extensions} to LLM
serving systems, executed during inference. We call these extensions
\emph{\MIapps}. An \MIapp implements its logic by monitoring and / or
modifying per-request model state, including tokens, hidden states, and the
residual stream. Multiple \MIapps can be deployed within \emph{the same}
inference engine instance, with a relevant subset activated for each request,
similar to multi-adapter serving systems~\cite{S-Lora,dlora}.

A unified \MIapp abstraction offers three key benefits. First, it encapsulates \MI logic and decouples it from the model, providing developers with a well-defined, easy-to-use interface. Second, it enables systematic runtime optimization of \MI execution. For example, an \MIapp may inspect hidden states, steer subsequent layers, or access intermediate activations that optimized GPU kernels never materialize. Executing such functions efficiently on a per-request basis requires tight integration with the serving runtime and compatibility with performance-critical optimizations such as continuous batching and CUDA graphs. Third, it enables \emph{system-level support} for deploying multiple \MIapps concurrently within a shared serving runtime, including  conflict-free state access, fast per-request activation and deactivation without service interruption, and KV-cache-aware \MIapp memory management.

Existing \MI frameworks fall short of supporting these requirements and incur prohibitively high runtime overheads. \TLENS{} and \NNSIGHT{} provide convenient access to internal model state~\cite{nanda2022transformerlens,fiottokaufman2024nnsight}, but suffer substantial slowdowns that preclude production use (\cref{SS:eval:OtherComp}). \ES is a steering-only framework integrated with \VLLM~\cite{xu2025easysteer}, but its optimized mode supports only a subset of steering applications, falling back to significantly slower execution otherwise (\cref{SS:eval:ESComp}). Importantly, none of these frameworks supports co-deployment and dynamic runtime toggling of multiple \MI functions, a key requirement for production serving.

We present \sys, a system that introduces the \MIapp abstraction as a first-class component of LLM inference serving. Developers implement \MIapps in Python using a simple domain-specific language (DSL) that specifies the submodules and layers whose state an app accesses, the computations it performs, and the conditions that trigger its execution. \sys compiles \MIapps into the model's execution flow within the original CUDA graph, ensuring conflict-free execution of multiple apps. Its runtime supports per-request activation and deactivation without service interruption, while dynamically reclaiming inactive \MIapp buffers for the KV cache.

Designing \sys requires addressing three key challenges.

\para{Ephemeral attention scores} Several important \MIapps~\cite{Charm,LLM-Check} operate on attention scores. However, high-performance inference frameworks rely on FlashAttention kernels~\cite{flashattention,FlashAttention-2}, which avoid materializing these scores, preventing their direct use by \MIapps. Naively recomputing the scores incurs substantial computation and memory overheads. Instead, we introduce \FI, a novel FlashAttention variant that exposes attention scores to \MIapps with low overhead, avoiding recomputation and reducing memory consumption (\cref{sec:FA}).

\para{\MIapp toggling} An \MIapp may comprise multiple GPU kernels. Consistent activation, deactivation, and memory management therefore require careful coordination across the serving engine's execution paths. We introduce an \MIapp registry that tracks deployed apps and their kernels, enabling efficient lifecycle management (\cref{SEC:runtime}).

\para{KV cache integration} Some \MIapps, particularly sparse autoencoders (SAEs), require large intermediate buffers. When deactivated, their memory should be reclaimed for the KV cache to increase request concurrency. However, CUDA graph compatibility requires preserving stable virtual memory addresses during serving. We leverage paged attention mechanisms to allocate KV cache pages to \MIapps, remap their underlying physical GPU memory, and return that memory to the KV cache upon deactivation.

We prototype \sys by integrating it with \VLLM and implement ten representative \MIapps across four state-of-the-art LLMs, demonstrating the generality and ease of use of our abstractions. Our evaluation shows low execution overhead across all applications, outperforming existing frameworks by up to two orders of magnitude. We further demonstrate fast per-request toggling of multiple co-deployed \MIapps without service interruption, efficient reclamation of inactive app memory to improve request concurrency, and no measurable overhead for deployed but inactive \MIapps.

\para{Contributions}
We make the following contributions:
{\setlength{\leftmargini}{1.2em}\setlength{\topsep}{2pt}\setlength{\partopsep}{0pt}\setlength{\parskip}{0pt}%
\begin{itemize}
    \setlength{\itemsep}{0pt}\setlength{\parsep}{0pt}%
    \item We introduce the \MIapp abstraction, guided by a comprehensive
    analysis of prior \MIMethods.
    \item We design a DSL, compiler, and runtime for efficient development, deployment, and execution of \MIapps.
    \item We introduce \FI, a novel low-overhead GPU kernel exposing ephemeral attention scores to \MIapps.
    \item We demonstrate that our \VLLM-based \sys prototype efficiently executes diverse \MIapps across state-of-the-art LLMs, supporting fast per-request toggling and GPU memory reclamation.
\end{itemize}}

\section{Background}
\label{SEC:background}

% Yaniv: this is not worth the space..backgroud is short and very browsable.
\ignore{
We first describe what are \MIapps
(\cref{SEC:background:apps}), then explain what are serving systems
and the optimizations they rely on
(\cref{SEC:background:serving_systems}), and finish with describing modern attention kernels (\cref{SEC:background:FlashAttention}).
}

%--------------------------------------------------------------------------
\subsection{Mechanistic Interpretability (MI)}
\label{SEC:background:apps}
\MI applications inspect or modify a model's internal computations at inference time, rather
than relying solely on its inputs and outputs. In an autoregressive
transformer, an \MI application may read hidden states, attention
values, or MLP activations at selected layers and token positions. It can
report a signal, such as a probe score, or write a modified activation, 
often called steering.

\MI intervention must be performed at the appropriate point in
the model's forward pass. It may read an activation at one layer and use
the result to modify a later layer during the same pass. For example,
conditional activation steering
tests a hidden state against a condition vector and, when the condition
holds, adds a refusal vector at a later layer~\cite{programmingrefusalconditionalactivation}. The serving system must
therefore preserve the order and location of \MI's read and write accesses.

Some applications can reuse activations produced during inference, avoiding a
separate model pass for the same signal. Their performance cost still
depends on how the serving engine exposes those activations, executes
the app's computation, and retains its state. Integrating these
operations into a high-throughput serving engine is the core
challenge addressed in this paper.

%--------------------------------------------------------------------------
\subsection{LLM Serving Systems and Existing Optimizations}
\label{SEC:background:serving_systems}
Production engines such as \VLLM~\cite{PagedAttention} and
SGLang~\cite{zheng2024sglangfull} attain high throughput through a set of
optimizations, each of which relies on an invariant about the forward pass. They maximize
throughput by applying multiple techniques. 
Below we briefly describe the ones relevant to this paper. 
%, such as disaggregated execution of prefill and
%decode, continuous batching, explicit KV cache management, and highly optimized GPU kernels such
%as FlashAttention \cite{flashattention}. They often rely on
%CUDA Graphs \cite{CudaGraphs} to reduce the overheads of recurrent kernel launches, %and introduce custom
%GPU kernels, including Triton-based fused kernels in vLLM, to keep performance-%critical operations
%on the device fast path \cite{tillet2019triton}. 

\para{Continuous batching}
Iteration-level batching~\cite{Orca} schedules a new batch at every forward pass, mixing prefill and decode tokens from different requests, allowing higher utilization of the GPUs compute resources. The engine
flattens them into a single tensor of shape
$[\textit{num\_tokens}, d_{\textit{model}}]$.

\para{CUDA graphs}
To eliminate the overhead of launching a GPU kernel from CPU, serving engines capture the forward pass as CUDA
graphs for a fixed set of padded batch sizes,  and replay
them at runtime without CPU involvement. Replay re-issues the recorded kernels with frozen tensor shapes and memory addresses.

\ignore{
\begin{figure}[t]
\centering
\TODO{Motivating figure. Panel (a): mean ITL / TTT for vLLM (CUDA graphs),
vLLM eager, and vLLM + PyTorch hook implementing the running example.
Panel (b): KV cache capacity and max concurrent requests vs.\ number of
deployed SAE layers. Optional panel (c): time to toggle an app via engine
restart.}
\caption{Cost of deploying \MIapps naively in \VLLM. \tbd{caption}}
\label{fig:background:motivation}
\end{figure}
}

%\subsection{FlashAttention}
%\label{SEC:background:FlashAttention}
\para{FlashAttention} Standard attention writes the
score matrix and its softmax to GPU memory and reads them back, resulting in space and I/O complexity 
quadratic in the sequence length. 
FlashAttention~\cite{flashattention} computes exact attention in a
fused kernel. It processes tiles sized to on-chip SRAM and
uses an online softmax that rescales partial results as each block of
keys arrives, so attention scores never reach GPU main memory. Since long prompts
and large batches make attention's cost dominant, FlashAttention-style
kernels are the default in \VLLM and SGLang.
%--------------------------------------------------------------------------

\section{The MI App Abstraction}
\label{SEC:abstraction}
\begin{table*}[t]
\centering
\tabletextsize
\setlength{\tabcolsep}{2.5pt}
\renewcommand{\arraystretch}{1.05}
\begin{tabular}{
    >{\raggedright\arraybackslash}p{0.40\textwidth}
    >{\raggedright\arraybackslash}p{0.12\textwidth}
    >{\raggedright\arraybackslash}p{0.10\textwidth}
    >{\raggedright\arraybackslash}p{0.10\textwidth}
    >{\raggedright\arraybackslash}p{0.20\textwidth}}
\toprule
\MIMethod & Purpose & Primitive & Condition & Submodule : Layers \\
\midrule
\textbf{1} Single Direction~\cite{arditi2024refusallanguagemodelsmediated}
& Safety & Write & \uncond & Post MLP : All \\

\textbf{2} SEAL~\cite{chen2025sealsteerablereasoningcalibration}
& Reasoning & Cond.~W & Token & Post MLP : All \\

\textbf{3} Cond.\ Refusal~\cite{programmingrefusalconditionalactivation}
& Safety & Cond.~W & Activations & Post MLP : Multi \\

\textbf{4} Hallucination Probe~\cite{orgad2025llmsknowshowintrinsic}
& Truthful. & Read & \uncond & Post MLP : Single \\

\ignore{Attention Hijack~\cite{bentov2025universal}
& Safety & Cond.~W & \red{TBD} & \red{TBD} \\
}

\textbf{5} SAKE~\cite{scialanga-etal-2025-sake}
& Knowledge & Write & \uncond & Post MLP : Single \\

\textbf{6} Side-Effect-Free Steering~\cite{stickland2024steering}
& Safety & Write & \uncond & Post MLP / Attn. : Multi \\

\ignore{FGAA~\cite{soo2025interpretable}
& \red{TBD} & Write & \red{TBD} & \red{TBD} \\
}

\ignore{SADI~\cite{wang2025sadi}
& Truthful. & Read $\rightarrow$ Write & \red{TBD} & \red{TBD} \\
}

\ignore{Representation Engineering~\cite{hojer2025representation}
& Reasoning & \red{TBD} & \red{TBD} & \red{TBD} \\
}

\ignore{Entity Awareness~\cite{ferrando2025entity}
& Truthful. & Read + Write & \red{TBD} & \red{TBD} \\
}

\ignore{FASB~\cite{cheng2025backtracking}
& \red{TBD} & Cond.~W & \red{TBD} & \red{TBD} \\
}

\textbf{7} Truthful Intervention~\cite{li2023inference}
& Truthful. & Write & \uncond & Post QKV : All \\

\textbf{8} Universal Prefixes~\cite{bentov2025universal}
& Jailbreak & Read & \uncond & Attn. scores : Multi \\

\textbf{9} SteerMoE~\cite{fayyaz2026steermoe}
& Safety & Cond.~W & Router & Router : All \\

\ignore{RICE~\cite{wang2025twoexperts}
& \red{TBD} & Write (Router) & \red{TBD} & \red{TBD} \\
}

\textbf{10} SAE~\cite{LlamaScope}
& {Feature ext.} & Read & \uncond & Post MLP : Single \\

\bottomrule
\end{tabular}
\vspace{-6pt}
\caption{These \RepAppCount \MIMethods represent
\TotalAppCount surveyed papers. \Cref{appx:survey} maps each paper to
its representative method.}
\label{tab:MIAPPSTab}
\end{table*}

\subsection{Survey of \MIMethods}
%\label{SS:abstraction:survey}
To inform the design of \sys, we surveyed \TotalAppCount prior \MIMethods
spanning safety, reasoning, truthfulness, knowledge, and feature extraction.
We analyzed the execution requirements for implementing each method as an
\MIapp in a serving system.
\cref{tab:MIAPPSTab} shows 
a representative subset, the full set appears in \cref{appx:survey}. We found that the interface between an app and the model is determined by the following properties: the app access \emph{primitive} to the model state, the \emph{condition} that triggers its execution, and the loci at which it accesses the model state as defined by 
\emph{submodule} and \emph{layers}. 

\para{Primitive} An app may read an intermediate value, write a
new value, or perform a conditional write. The hallucination probe
(\#4) and sparse-autoencoder feature extractor (\#10) are
Reads; Single Direction (\#1) and SAKE (\#5) are Writes.
Conditional Refusal (\#3) and SteerMoE (\#9) each evaluate a
condition before writing. This classification specifies the app's
effect on model state without prescribing what computation is performed on it.

\ignore{
\para{Condition} The condition states when an app acts. The table includes
unconditional execution (\#1), a token trigger (SEAL, \#2), an
activation trigger (Conditional Refusal, \#3), a router trigger
(SteerMoE, \#7), and a last-token trigger (Universal Prefixes,
\#10). These are distinct conditions even when apps share a
primitive: \#2, \#3, and \#7 are all conditional Writes, but
their conditions arise from different parts of inference. Selection of
which installed apps a request uses is separate from these triggers.
}

\para{Condition} The condition states when an app should be invoked while processing a request . 
 Some surveyed apps are invoked unconditionally (\#1), others are conditioned on signals obtained from activations (\#3), 
some others are triggered only when a specific set of tokens appear (\#2). The condition is not to be confused with the selection of which apps act on 
which request.

\para{Submodule} The submodule identifies the value an app accesses within a model layer.  
Most apps operate after the MLP (see \cref{tab:MIAPPSTab}), but other cases are common too. App~\#8
reads \FI's thresholded (query, key, score) list for jailbreak detection;
position-specific analysis and prevention are not evaluated. Thus, we support
target submodules that reside before, after, and within attention
and MLP blocks, spanning: residual additions, normalization layers, projection 
inputs/outputs, router logits, and activation functions.
\sys offers eight submodule hook locations, which we found sufficient 
to cover all surveyed apps. Full list is provided in 
\cref{appx:loci}.

\para{Layers} The layer identifies whether an app runs at one or
several specific layers, or all layers. The hallucination probe (\#4)
reads one layer, Conditional Refusal (\#3) writes at multiple
layers, and Single Direction (\#1) writes at all layers. Submodule
and layer selection are independent: an app can target a different
submodule without changing how many layers it covers, or change its
layer coverage while keeping the same submodule.

% Yaniv: I moved this to comment as it doesn't say anything not said in the next subsection
%\para{The contract}
%An app's execution requirements map to the four properties described above, defining its contract, regardless of the end goal it aims to achieve. For example, both (\#1) and (\#7) are used to ensure safety, yet vary greatly by their contract. The \sys \DSL (see \cref{SS:abstraction:DSL}) describes the \MI logic contract, which once compiled, becomes an MI app.

\begin{figure}[]
  \centering
\begin{codebox}[natural height, top=0pt, bottom=0pt]
\begin{lstlisting}[
  numbers=left,
  numberstyle=\tiny\color{black!60},
  numbersep=6pt,
  xleftmargin=2em,
  escapeinside={(*@}{@*)},
  basicstyle=\ttfamily\fontsize{8pt}{9pt}\selectfont
]
Import SteerLogic, CosSim
steer = SteerLogic(refusal_vec)(*@\label{line:cast-dsl:steer}@*)
cossim = CosSim(condition_vec, (*@$\theta$@*))(*@\label{line:cast-dsl:condition-def}@*)
App3 = (m.write(steer.run)     (*@\label{line:cast-dsl:writeprim}@*)
  .layer[15:23](*@\label{line:cast-dsl:write-layer}@*)
  .submodule("mlp.post")(*@\label{line:cast-dsl:write-submodule}@*)
  .cond(cossim.layer([12])(*@\label{line:cast-dsl:condition}@*)
    .submodule("mlp.post")))(*@\label{line:cast-dsl:condition-submodule}@*)
\end{lstlisting}
\end{codebox}
\vspace{-8pt}
  \caption{App \#3, an example of \sys's DSL.}
  \label{fig:cast-dsl}
\end{figure}

%The interface above serves the basis for the \sys's DSL explained below. 
\subsection{\sys's Domain Specific Language}
%\label{SS:abstraction:DSL}
\sys developers express \MI logic in \sys's \DSL by specifying the four properties described above. 
\Cref{fig:cast-dsl} shows the implementation of the 
Conditional Refusal \MIapp (\#3) we use as a running example.
This app reads the hidden state at layer $12$, projects it onto a condition vector, 
and, if the similarity exceeds a threshold $\theta$, adds a refusal vector at later
layers $15$-$23$.

Line 1 in \cref{fig:cast-dsl} imports \scode{SteerLogic} and
\scode{CosSim} from \sys's kernel library
(\cref{sec:design:compiler:kernel_library}). The kernels implement the
computations on the state; the \DSL specifies where and when they run.
\cref{line:cast-dsl:steer} initializes the steering logic with the
refusal vector. \cref{line:cast-dsl:condition-def} constructs the
\scode{CosSim} condition from the condition vector and threshold $\theta$. On
\cref{line:cast-dsl:writeprim},
\scode{m.write(steer.run)} selects the Write primitive and its steering
kernel. The write layers (\scode{.layer[15:23]}) are specified on
\cref{line:cast-dsl:write-layer}, and the write submodule
(\scode{.submodule("mlp.post")}) on
\cref{line:cast-dsl:write-submodule}. On \cref{line:cast-dsl:condition},
\scode{.cond} attaches the condition and \scode{.layer([12])} places its
evaluation at layer 12. \cref{line:cast-dsl:condition-submodule} places the
condition after the MLP block through \scode{.submodule("mlp.post")}.

\section{Design}
\label{SEC:design}

\sys makes \MIapps first-class entities of the serving system. Before
serving begins, its compiler lowers each app's \DSL into serving-path
operations at the requested model locations and gives the runtime an
identity and resources for each app (\cref{SS:design:compiler}). \FI
exposes attention scores to apps
(\cref{sec:FA}). During serving, the runtime selects apps per request,
activates or deactivates them, and returns inactive app memory to the
KV cache (\cref{SEC:runtime}).

\subsection{The Compiler}
\label{SS:design:compiler}

The compiler turns an ordered list of \MIapps into an execution path that the serving engine can capture and replay. It receives their \DSL implementations and registered kernels, plus an engine-specific map from model locations to code insertion points. For apps targeting the same layer, list order determines access order. The compiler produces integrated serving-engine code and a registry that identifies each app and its resources to the \sys runtime, in two stages: code generation and integration.

\para{Code generation} \sys applies a predefined lowering rule to each \DSL statement and collects the resulting Python snippets in an ordered map keyed by insertion point. The map groups snippets by template placeholders for app construction, buffer setup, and kernel calls. For a Read or Write, the submodule selects the insertion point within a layer, and the layer selector determines which layers receive a kernel call. A Conditional Write generates a predicate call and one or more subsequent Write calls that exchange data through the app’s GPU scratch space. The compiler also emits named kernel registrations and assigns each app an ID for the runtime’s per-request bitmask (\cref{SEC:runtime}).

\para{Application structure}\label{SEC:compiler:template} Generated calls use kernels built on the \sys base class {\texttt{MI\_base}}, which defines a per-app GPU scratch space shared among all of that app’s kernels. Each app uses fixed-capacity input and output buffers, and its kernels are implemented in Triton. Thanks to the shared scratch space, a result computed at one model location can guide a later kernel without a host transfer. For example, the \scode{CosSim} kernel (\cref{fig:cast-dsl}) writes a per-token predicate into scratch space at layer 12, and \scode{SteerLogic} reads it to decide whether to apply the refusal vector at layers 15–23. This coordination realizes Conditional Write, but it is internal to the kernels and invisible to the DSL.
%while keeping per-app state isolated.

\para{Kernel library}\label{sec:design:compiler:kernel_library} \sys ships 15 kernels covering the surveyed applications, all built on {\texttt{MI\_base}}. Developers can also register custom kernels. All kernels avoid host synchronization, so the serving engine can capture and replay them in CUDA graphs as part of the model.

\para{Integration} The engine-specific map resolves each logical insertion point to a predefined placeholder in the serving-engine template. \sys fills those placeholders with the collected snippets in app order: construction snippets instantiate each app and its buffers when the model is built; buffer-setup snippets connect each app’s kernels to its own scratch space; and kernel-call snippets run at the selected points in the forward pass. The result keeps all GPU operations in the model execution path while allowing the runtime to manage each app independently.

\para{Correctness} During code generation, \sys type-checks every kernel-call argument against the registered kernel’s interface, and rejects statements that name an unregistered kernel, omit a required argument, or target a submodule the model does not expose. It also rejects a Conditional Write whose predicate is not computed at a layer that strictly precedes the Write. During integration, each app receives its own scratch space and buffers, isolating app state even when multiple apps execute at the same model location.
\subsection{FlashInspect}
\label{sec:FA}

Every submodule but one is a value the forward pass already writes to
memory, so the compiler only needs to install applications there. The
exception is attention scores, which several applications
consume~\cite{Charm,bentov2025universal,Lookback_Lens,LLM-Check} but
\ac{FA} never materializes. Instead of falling back to less efficient attention 
implementations, when an \MIapp consumes attention
scores, the compiler replaces the \ac{FA} kernel with our \FI
that also exports them.

Below we reiterate the \ac{FA} algorithm and describe the modifications we made 
to create \FI. 

% =====================================================================
%  FlashAttention-2 forward pass + threshold retention.
%  Written in `algorithmic' syntax (ALL-CAPS commands) because
%  mlsys2025.sty already does \RequirePackage{algorithmic}.
%  Do NOT load algpseudocode -- that is what broke the line numbers.
%
%  Add to your preamble only what is missing:
%      \usepackage{amsmath,amssymb}
%      \usepackage{float}       % only for [H]; use [tb] to omit it
%      \usepackage{setspace}    % only for \setstretch
% =====================================================================
% Diff-style marking of lines added to the original FlashAttention-2:
% a light green band behind the line's text and a green "+" just left of
% the algorithm box, aligned across nesting levels. Neither affects layout.
\definecolor{diffaddbg}{RGB}{220,255,220}
\definecolor{diffaddfg}{RGB}{26,127,55}
\makeatletter
\newcommand{\faadded}[1]{%
  \makebox[0pt][l]{\hspace{-0.3em}%
    \textcolor{diffaddbg}{\rule[-0.45em]{\dimexpr\linewidth+0.3em\relax}{1.55em}}}%
  \makebox[0pt][r]{\textcolor{diffaddfg}{\textbf{+}}%
    \hspace{\dimexpr\@totalleftmargin+0.4em\relax}}%
  #1}
% algorithmic does not \refstepcounter its line counter, so a plain \label
% would not resolve to the line number.
\newcommand{\falabel}[1]{%
  \protected@edef\@currentlabel{\arabic{ALC@line}}\label{#1}}
\makeatother
% Tighter space between the algorithm and the surrounding text.
\begingroup
\setlength{\intextsep}{4pt plus 1pt minus 1pt}
\begin{algorithm}[H]
\caption{\textsc{FlashInspect} forward pass with threshold retention
         (one query block $\mathbf{Q}_i$; $\mathcal{M}$ = retained candidates,
         $\ell$ = running rowsum, $m$ = running rowmax)}
\label{alg:fa2-threshold}
\small                   % caption stays at normal size: these come after it
\setstretch{1.25}        % <-- ROW SPACING. 1.0 = default, 1.5 = airy.
\begin{algorithmic}[1]
\REQUIRE $\mathbf{Q}_i$; blocks $\mathbf{K}_j,\mathbf{V}_j$, $j=1,\dots,T_c$;
         threshold $\tau$.
\STATE $\mathbf{O}^{(0)}\!\gets\!0$, $\ell^{(0)}\!\gets\!0$,
       $m^{(0)}\!\gets\!-\infty$, $\mathcal{M}\gets\emptyset$
\FOR{$j=1,\dots,T_c$}
\STATE\falabel{alg:fa:scores} $\mathbf{S}^{(j)}\gets\tfrac{1}{\sqrt{d}}\,\mathbf{Q}_i\mathbf{K}_j^{\top}$
\STATE\falabel{alg:fa:max} $m^{(j)}\gets\max\!\big(m^{(j-1)},\,\mathrm{rowmax}\,\mathbf{S}^{(j)}\big)$
\STATE\falabel{alg:fa:alpha} $\alpha^{(j)}\gets e^{\,m^{(j-1)}-m^{(j)}}$
\STATE $\tilde{\mathbf{P}}^{(j)}\gets\exp\!\big(\mathbf{S}^{(j)}-m^{(j)}\big)$
\STATE\falabel{alg:fa:sum} $\ell^{(j)}\gets\alpha^{(j)}\ell^{(j-1)}+\mathrm{rowsum}\,\tilde{\mathbf{P}}^{(j)}$
\STATE\falabel{alg:fa:out} $\mathbf{O}^{(j)}\gets\mathrm{diag}\big(\alpha^{(j)}\big)\mathbf{O}^{(j-1)}
       +\tilde{\mathbf{P}}^{(j)}\mathbf{V}_j$
\STATE\falabel{alg:fa:retain} \faadded{$\mathcal{M}\gets\mathcal{M}\cup
       \big\{(r,c,S^{(j)}_{rc}):\tilde{P}^{(j)}_{rc}\ge\tau\,\ell^{(j)}_r\big\}$}
\ENDFOR
\STATE $\mathbf{O}_i\gets\mathrm{diag}\big(\ell^{(T_c)}\big)^{-1}\mathbf{O}^{(T_c)}$
\STATE\falabel{alg:fa:lse} $L_i\gets m^{(T_c)}+\log\ell^{(T_c)}$
\STATE\falabel{alg:fa:compact} \faadded{$\mathcal{M}\gets\big\{(r,c,p):(r,c,s)\in\mathcal{M},\,
       p=e^{\,s-L_{i,r}}\ge\tau\big\}$}
\STATE \textbf{return} $\mathbf{O}_i$, $L_i$, $\mathcal{M}$
\end{algorithmic}
\end{algorithm}
\endgroup

\para{Vanilla attention scores computation} In the vanilla version (before \FA), 
to compute each 
attention head, every token distributes
its attention over the preceding tokens. Token representations are
projected into queries $Q$, keys $K$, and values $V$, and the head
computes $\mathrm{Softmax}(QK^\top/\sqrt{d_k})\,V$. The softmax output
$P$ holds the attention scores: $P_{r,c}$ is the weight token $r$
places on token $c$.

\para{Algorithm} \cref{alg:fa2-threshold} shows \FI
forward pass for one query block, as a diff on top of the 
\ac{FA} computation. Unlike the vanilla version, \ac{FA} never forms $P$. 
For a query row, it
iterates over blocks of keys; at block $j$ it computes the scores
$S^{(j)}$ (line~\ref{alg:fa:scores}) and updates a running maximum
$m^{(j)}$ (line~\ref{alg:fa:max}) and running sum $\ell^{(j)}$
(line~\ref{alg:fa:sum}) of the exponentiated scores, rescaling earlier
partial results as $m$ grows (lines~\ref{alg:fa:alpha},
\ref{alg:fa:sum}--\ref{alg:fa:out}). The final score of key $c$ is
$e^{\,S_c - L}$, where $L = m + \log \ell$ (line~\ref{alg:fa:lse}) is
known only after the last block. By then, each block's partial scores
have been multiplied into the output (line~\ref{alg:fa:out}) and
discarded. Even at the end, \ac{FA} holds only their product with $V$,
never the attention scores themselves.

\para{Insight} Applications rarely need all of $P$: they read only the
scores above a threshold $\tau$~\cite{Charm} or at specific
positions~\cite{bentov2025universal,LLM-Check}. Materializing $P$ would
take memory quadratic in the context length from the KV cache and
forfeit the I/O savings of \ac{FA}. 
Instead, we designed \FI. \FI
iterates over key blocks, it retains every key whose partial score
$\tilde{P}^{(j)}_{c}/\ell^{(j)}$ exceeds $\tau$ (Line~\ref{alg:fa:retain}). 
Since $\ell$ only grows,
a key's partial score only decreases. The retained keys therefore form a
superset of those whose final score exceeds $\tau$. Once the attention
kernel completes and $L$ is known, a separate compaction kernel computes
the final scores of the retained keys and discards those below $\tau$ (line~\ref{alg:fa:compact}).

\para{Output and memory} This computation outputs a buffer that contains a list of tuples 
(attention scores, token, head, layer), only for those whose attention score exceeds $\tau$. Thus, applications consume this sparse list rather than dense activations.
A row's final scores sum to one, so at most $\lfloor 1/\tau\rfloor$ tuples
exceed $\tau$. This allows us to bound the size of this buffer. 
Per head and layer, the output therefore holds at most
$N\lfloor 1/\tau\rfloor$ entries instead of the $N^2$ of $P$.
The candidate set before compaction must be larger (we use $16\times$ of  $\lfloor 1/\tau\rfloor$) , since (line ~\ref{alg:fa:retain}) covers only the keys seen so far. %Hence, we pad the candidate buffer to the nearest KB, per row, head, and layer. 
If, however, the buffer capacity is insufficient, the kernel sets flag to app’s output so that the output for that request is marked incomplete rather than silently truncated. However, in our experiments we never saw the flag being set.
\subsection{The Runtime}
\label{SEC:runtime}

\begin{figure}[t]
  \centering
  \includegraphics[width=\columnwidth]{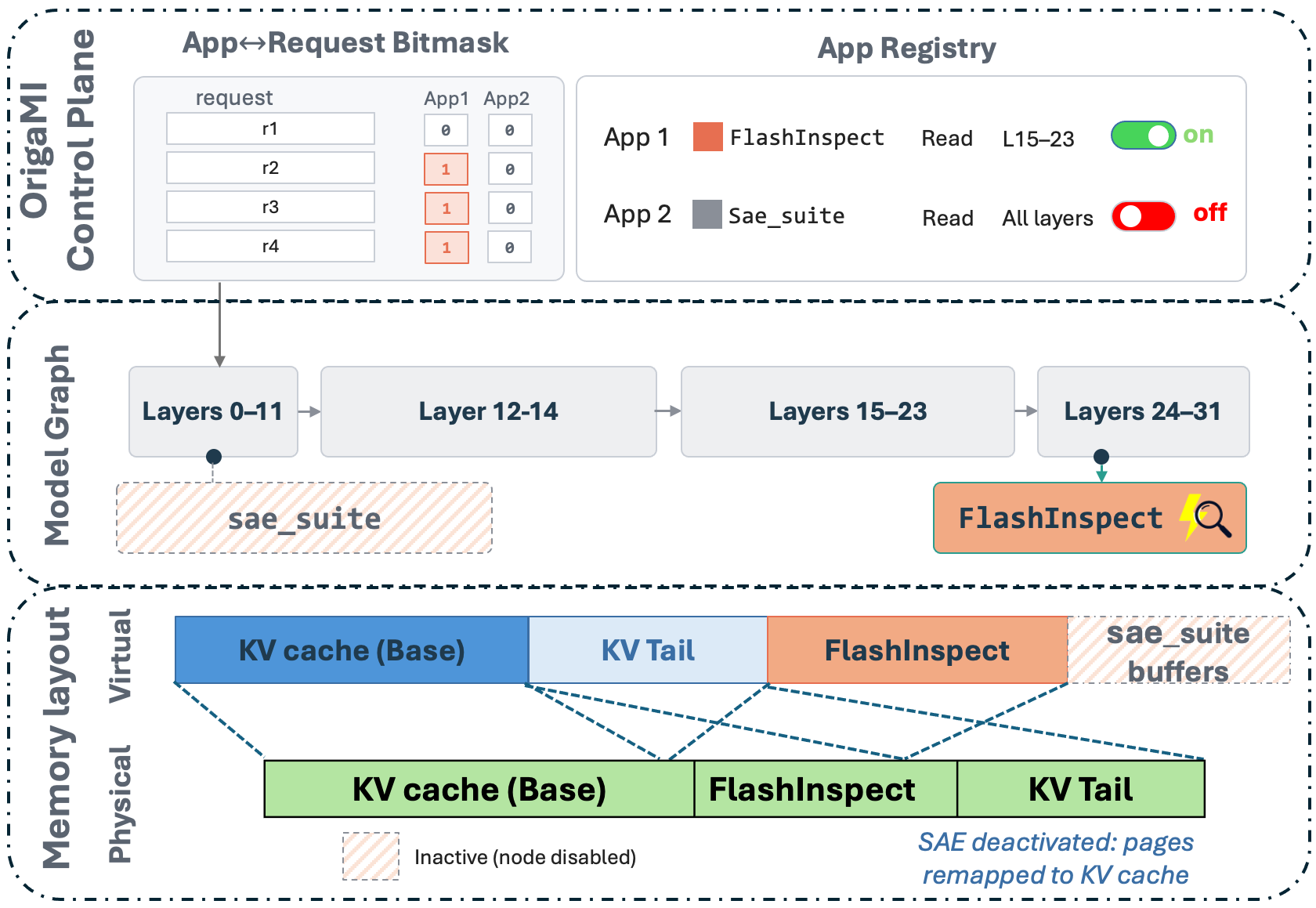}
  \caption{The \sys runtime with Universal Prefixes (App~8) enabled
  for requests r2-r4, and \scode{SAE-MultiLayerSuite} (App~10)
  deactivated. At layers 24 through 31, \FI replaces the FA
  kernel and supplies scores to app 8's kernel.
  The hatched SAE node represents the suite's disabled nodes across
  layers; its buffer pages are remapped to the KV cache tails.}
  \label{fig:runtime}
\end{figure}

The compiler leaves the calls forming the \MIapp in the model's execution path
and gives the runtime each \MIapp's identity and resources. As
\cref{fig:runtime} shows, \sys's runtime has three parts: (1) a control plane
with an \MIapp registry, application toggling, and a shared request
bitmask; (2) model graph containing the \MIapp kernels at their chosen locations; (3)
a memory layout that preserves virtual addresses while assigning
physical pages to application buffers or the KV cache. These structures
let \sys manage \MIapps during live serving.

\para{MI apps as CUDA graph objects} At engine initialization, the
serving engine captures the integrated forward pass as CUDA graphs.
As shown in \cref{fig:runtime}'s middle bar, each application's kernel launches become 
ordinary graph nodes, and the compiler's named kernel registrations let \sys 
associate those nodes with the application. For example, the \FI kernel (on the right) 
supplies attention scores to the Universal Prefixes app kernel. 
Request-specific conditions are evaluated inside the kernels
through the shared bitmask, rather than through CUDA conditional nodes.

\para{App selection} As \cref{fig:runtime}'s upper bar shows,
each request may exclude active
applications through its configuration. The runtime records these
choices in a single bitmask buffer shared by all applications, holding
one bit per request-application pair: a set bit allows the application
to act on that request. At every iteration, the runtime fills the
bitmask from the batch's composition. Each application's kernel reads
its own bits, located by the ID the compiler assigned it, to determine
which rows of the batch it acts on.
Because the bitmask lives in a fixed buffer, this requires no change to
the graph. Applications thus apply to exactly the right rows of a
continuously changing batch that all requests share.

\para{MI app registry} An operator toggles an application
through the serving interface. To deactivate it, the runtime disables its
nodes in the captured graphs through the CUDA driver API, between
iterations. Its kernels are not launched, effectively eliminating
deactivated application invocation cost. Reactivation re-enables the
same nodes. Under tensor parallelism, the runtime applies each toggle on
all ranks, so their graphs remain identical.

\ignore{
For comparison, whereas kernel toggling takes micro-seconds, engine restart takes tens 
of seconds \cite{foundry}, several orders of magnitude larger than a 
whole forward pass.}

\para{Memory reclamation} As shown in \cref{fig:runtime}'s lower bar, 
application deactivation also returns its allocated
memory to the KV cache. For applications with large footprints,
this is substantial: a \ac{SAE} suite occupies {11.5\,GB}. 
%As 
%\cref{fig:eval:reclaim-throughput} shows,
%reclaiming memory of already inactive applications increases KV cache 
%capacity by {50\%}, raising throughput by {20\%}.
%
The difficulty is that captured graphs hold fixed addresses: those of
application buffers, and the base address of each layer's KV cache,
from which the attention kernel offsets into blocks through the block
table. Memory can therefore move between them only beneath those
addresses. \sys separates addresses from physical memory using CUDA
virtual memory management. Each layer's KV cache reserves a virtual
range consisting of a base, mapped at startup, and a tail, mapped only
when memory is freed, so new blocks lie within the range the attention
kernel already indexes. 

When an \MIapp is deactivated, the runtime
unmaps its buffers' physical memory, swaps their content to the CPU, and maps it into the KV cache
tails. Since a block spans all layers, the memory is divided equally
among the layers, and the engine adds the resulting blocks to its
free pool, where they serve new requests. 

\ignore{
Upon reactivation, \sys reclaims memory using an efficient preemption algorithm. To minimize recompute preempted requests overheads, it preempts the younger ones first until enough physical memory is freed. Crucially, \sys recompacts the KV cache of surviving requests to prevent block fragmentation. The freed pages are then mapped back under the application's fixed virtual buffers and restored from the CPU, leaving captured graph addresses unchanged.}

Upon reactivation, \sys reclaims memory using an efficient preemption algorithm. To minimize recompute preempted requests overheads, it preempts the younger ones first until enough physical memory is freed. Crucially, \sys recompacts the KV cache of surviving requests to prevent block fragmentation. The freed pages are mapped back under the application's fixed virtual buffers and restored from the CPU, leaving captured graph addresses unchanged.

\ignore{
When the application is reactivated, the runtime
reclaims the same amount of memory from the tails: it unmaps tail
blocks, preempting the requests that hold them (vLLM
later rebuilds their KV cache by re-prefilling the prompt and the tokens
generated so far), and maps the reclaimed
memory back under the application's buffers, copying the swapped buffer contents. Neither direction
changes an address a captured graph uses.
}
%\TODO{Elaborating about re-activation scenario}

% YANIV: I didn't see any reason to go into these last details.
% Also, the last two paragraphs seem a bit too detailed for this section, and could be moved to an appendix.

\para{Output path} An application's outputs stay on the GPU by design:  
Signals consumed outside the forward pass,
such as probe scores or SAE feature indices, are typically a few bytes per request and
are copied device-to-host asynchronously at the iteration boundary, where the
engine already synchronizes with the CPU for scheduling.

%a signal used as a predicate (Fig.~1) is consumed by the application's own
%kernels, with no host round trip.

\section{Evaluation}
\label{SEC:eval}

\begin{figure*}[t]
  \centering
  \includegraphics[width=\textwidth]
    {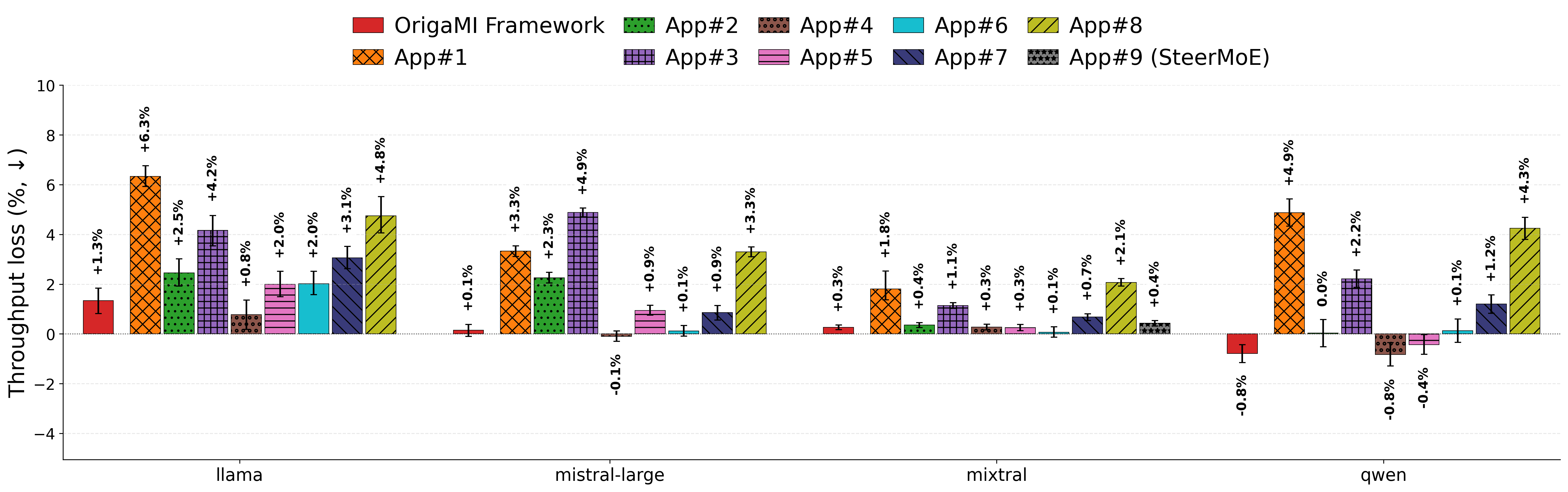}
  \caption{\acs{TTT} loss relative to the \VLLM baseline for Apps \#1--\#9,
  deployed with \sys. Lower is better.
  Each group reports a per-model breakdown for Llama, Mistral,
  Mixtral, and Qwen3. Error bars denote 95\% confidence intervals.}
  %The remaining
  %metrics are reported in \cref{appx:eval-%metrics}. 
  
  \label{fig:eval:apps-ttt}
\end{figure*}

We evaluate \sys to demonstrate that it fulfills all our design requirements across the following axes:
%\yanivRes{Questions needs to be sharper+shorter. Cost is confusing (its not \$ cost..its runtime overhead...). Also, frame positively, e.g., "How efficient is OrigaMI is serving MIApp mixtures.."}
{\setlength{\topsep}{2pt}\setlength{\partopsep}{0pt}\setlength{\parskip}{0pt}%
\begin{itemize}
    \setlength{\itemsep}{1pt}\setlength{\parsep}{0pt}\setlength{\parskip}{0pt}%
    \item \textbf{Flexibility:} \sys abstractions are sufficient to implement ten state-of-the-art representative \MIapps including SAE, steering, truthfulness and jailbreak detection (\cref{tab:MIAPPSTab}).
    \item \textbf{Low overhead:} \sys is efficient across all implemented \MIapps and frontier LLMs of different sizes, consistently reducing the overheads of the state-of-the-art alternatives by one or two orders of magnitude (\cref{SEC:eval:cost}).
    \item \textbf{\FI:} \FI enables low-overhead implementation of \MIapps that require attention scores, incurring $4.5\times$ lower overheads than score recomputation
    (\cref{SEC:eval:fa}).

    \item \textbf{Toggling:} \sys toggling achieves fast switching of deployed \MIapps without interrupting inference, three orders of magnitude faster than restarting the model to disable an app, and incurs no measurable slowdown for deployed but inactive \MIapps (\cref{SEC:eval:runtime}).
    
    \item \textbf{SAE:} 
     \sys's memory reclamation mechanism enables full throughput when dynamically toggling of SAE kernels by replenishing the KV-cache memory to achieve maximum request concurrency (\cref{SEC:eval:sae}). 
    
\end{itemize}}

%--------------------------------------------------------------------------
\subsection{Experimental setup}
\label{SEC:eval:setup}
\para{Applications and models} We evaluate \scode{Apps \#1-\#9} , the
representative applications of \cref{tab:MIAPPSTab}, implemented using \sys's \DSL (\cref{SEC:design}). Together they
cover all three primitives, all condition types, and single-layer,
multi-layer, and all-layer selection options. \scode{Apps \#1-\#8} run on
Llama-3.1-8B-Instruct, Qwen3-8B, Mixtral-8x7B-Instruct-v0.1, and Mistral-Large-Instruct-123B. The
\ac{MoE}-specific \scode{App \#9} runs on Mixtral only. We also implement and evaluate SAEs (\scode{App \#10}, \cref{SEC:eval:sae}).

\para{Workload} We use \VLLMNS's
\scode{bench sweep} utility~\citep{vLLM-Bench-sweep-serve}. Each run
issues 250 prompts sampled from ShareGPT~V3~\citep{ShareGPT} under an unconstrained arrival rate. This saturates the engine's admission queue and maintains maximum batch occupancy, evaluating the system under peak throughput conditions. For models
without published steering vectors, we use random vectors of matching
dimension. The output length is fixed, thus the content vector does not change the
work performed.

\para{Baseline} Our baseline is vanilla \scode{vLLM 0.11.1} 
%with \sys's hooks \yanivRes{this is platform only? if it is the full explanation should go to an appendix}
%\michael{No, its real vanilla vllm}
%\yanivRes{So, say why not say vanilla vLLM? hooks uninstalled seem like you added code but then did some things to disable it }
%uninstalled. %\TODO{Either address the fact that the platform itself is also measured, or omit it} 
%\ms{I actually don't understand what "hooks uninstalled" means. It's either all hooks not populated, or vanilla vLLM. }
%All comparisons use \VLLM's
%default configuration, including CUDA graphs, chunked prefill, and
%prefix caching.
%
All comparisons use \VLLM's
default configuration, including CUDA graphs and chunked prefill.

\para{Hardware} We use AMD EPYC 7742 64-core CPU and
NVLink-connected A100-SXM4-80GB GPUs. All experiments run \VLLM with
tensor parallelism on two GPUs, except for Mistral-Large-Instruct-123B on eight. 

\para{Correctness} Without steering \MIapps, \sys's outputs are
token-for-token identical to vanilla \VLLM. Where the true steering
vector is available, \eg Llama-3.1-8B-Instruct's refusal vector of \scode{App~\#1}
\citep{arditi2024refusallanguagemodelsmediated}, we verified that steering produces the intended
behavior.

\para{Methodology} For each configuration, we discard the first three
runs as warmup and aggregate the remaining seven. We report the
arithmetic mean for latency and the harmonic mean for normalized throughput vs. the baseline, with 95\% confidence intervals.

%\TODO{Should we state the entire span, consider rephrasing}

%--------------------------------------------------------------------------

\subsection{Single-application performance}

We measure \sys's performance across four models and nine \MIapps, each run alone.
We measure \ac{TTT}, \ac{ITL}, P99 \ac{ITL}, and \ac{TTFT}.
Across all these metrics the results follow the same trend, so here we report only \ac{TTT} and show the rest in \cref{appx:eval-metrics} due to the lack of space.

\ignore{
We measure \sys's performance across four models and nine \MIapps when running each \MIapp alone. 
We measure four key metrics:  \ac{TTT}, \ac{ITL}, P99 \ac{ITL}, and \ac{TTFT}.
Across all these metrics the results follow the same trend, so here we report only \ac{TTT} and show the rest in \cref{appx:eval-metrics} due to the lack of space.}

\Cref{fig:eval:apps-ttt} shows the slowdown compared to vanilla \VLLM. Input/output length fixed at 512/128 tokens respectively. 
The first bar is the overhead of \sys without \MIapps. 

%At \cref{fig:eval:apps-ttt} we 

%report \TTT, which reflects serving capacity. 
%Mean \ITL, which reflects the delay added to every generated token.
%Tail (P99) \ITL and \TTFT follow the same trends; all four metrics for
%every application and model are reported in . \yaniv{This appendix reference should move %to where the partial results are presented. }
%\michael{It seems like we hide it, we don't, it just doesn't fit in eval., I also %reference it there}
%\yanivRes{You should not point to the figure at all, or the appendix. In the setup you just say what you are measuring..}
%\yanivRes{when toggelled on? (to stress its running alone)}
%
\emph{We observe low overhead for all evaluated cases, with the harmonic mean of 1.84\%.}
As expected, the differences in performance are app-dependent, and relative performance across \MIapps is consistent across models. 
Each \MIapp adds a different amount of work to the forward pass, so the overheads correlate with the app's compute load. \scode{App-1} manipulates all model layers,
and thus incurs the highest overhead.
\scode{App-4} reads hidden states on a single layer and runs a probe, so it is among the cheapest. 
\scode{App-3}  writes on a quarter of the layers, and is conditioned on a predicate, so it falls in
between.
Unexpectedly, Qwen runs even slightly faster with some apps than without the baseline. We traced it to memory alignment caused by change in memory layout due to additional code and buffers.

\begin{comment}
    
\para{Model architecture and size} 
We observe that the overheads do not grow with the model size.
On Mixtral, the performance is affected by at most 
{1.8\%} across all applications, including the \ac{MoE}-specific \scode{App-7}, which
intervenes on router logits. On Mistral-Large-Instruct-123B across
eight GPUs, overheads remain below {4.9\%} (\cref{fig:eval:apps-ttt}).
Larger models like Mistral, have more layers and a larger hidden dimension, this extra cost is balanced by the proportionally larger forward pass, keeping relative overheads roughly flat rather than compounding with scale.
\end{comment}

%--------------------------------------------------------------------------
\subsection{\FI for attention scores}
\label{SEC:eval:fa}

\begin{figure}[t]
  \centering
  \begin{subfigure}[t]{0.45\columnwidth}
    \centering
    \includegraphics[width=\columnwidth]
      {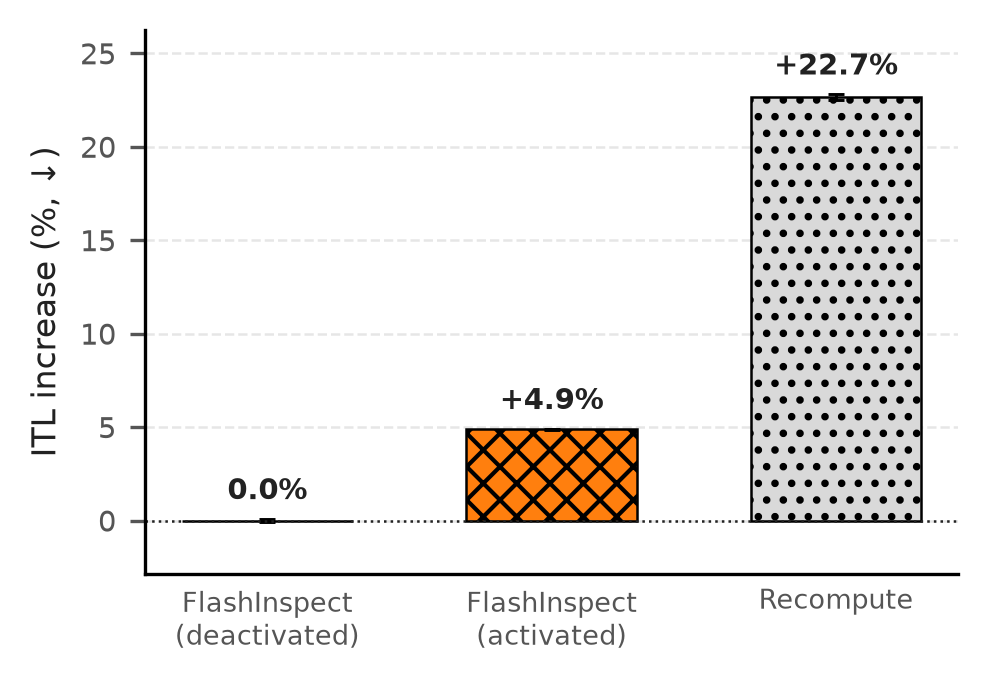}
    \caption{Mean \acs{ITL}}
    \label{fig:eval:fa-itl}
  \end{subfigure}\hfill
  \begin{subfigure}[t]{0.45\columnwidth}
    \centering
    \includegraphics[width=\columnwidth]
      {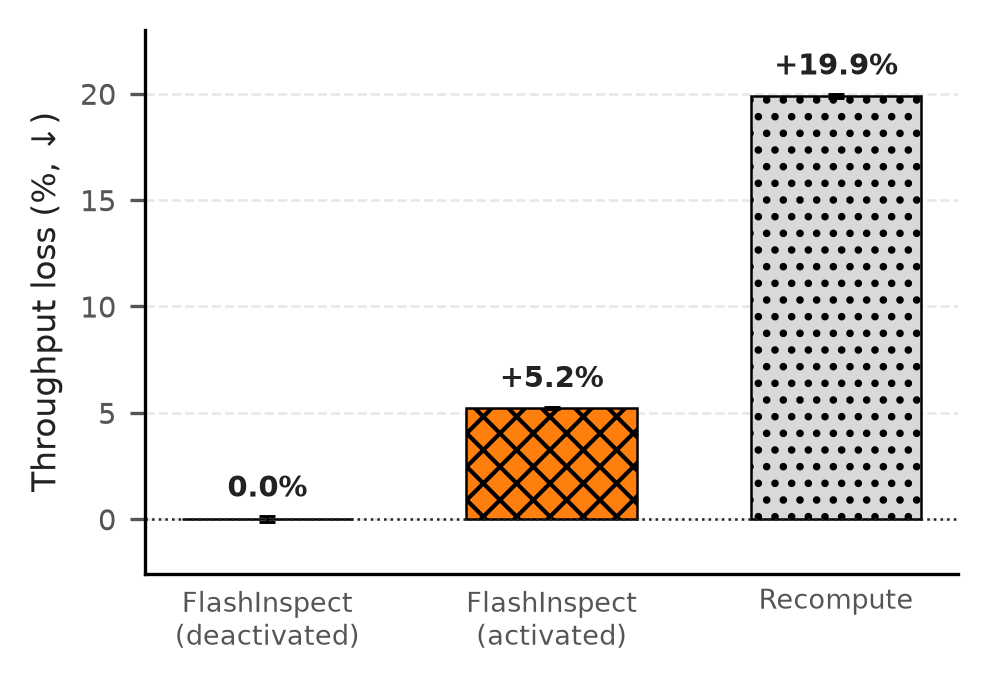}
    \caption{\acs{TTT}}
    \label{fig:eval:fa-ttt}
  \end{subfigure}
  \vspace{-6pt}
  \caption{Overheads of \FI over FlashAttention for exposing attention scores on Llama-3.1-8B-Instruct on long context queries. Lower is better. }
  \label{fig:eval:fa}
\end{figure}

\Cref{fig:eval:fa} shows the performance of using \FI for \MIapps that rely on attention scores. We evaluate on  Llama-3.1-8b-Instruct with input/output length fixed at 16K/2k tokens respectively.

When a deployed \MIapp is inactive, \FI and FlashAttention are equivalent.
With no \MIapps deployed, the original FlashAttention kernel runs (\cref{sec:FA}).

\ignore{
When a \MIapp is deployed but inactive,  \FI and FlashAttention are equivalent.
Note that when no \MIapps are deployed, the original FlashAttention kernel is invoked (\cref{sec:FA}).
}

Extracting the
scores via \FI increases mean \ITL by 4.9\% and
reduces \TTT by 5.2\%. However, these overheads are about $4.5\times$ lower compared to 22.7\% and 19.9\% for
recomputation (\cref{fig:eval:fa}). 
Moreover, \FI retains only attention scores exceeding a threshold $\tau$,
keeping memory bounded by $N\lfloor 1/\tau \rfloor$ entries per head and
layer for 
a sequence of length $N$. At most $\lfloor 1/\tau \rfloor$ entries per query regardless of the
context length.
%--------------------------------------------------------------------------

%--------------------------------------------------------------------------

%--------------------------------------------------------------------------

\subsection{Toggling \MIapps at runtime}
\label{SEC:eval:runtime}

\para{Toggle latency} Without \sys, enabling or disabling an
application can be done in two ways: by  restarting the \VLLM service to reload the CUDA graph; or by using an eager mode without CUDA graphs. 

Both approaches incur prohibitively high performance costs.

Whereas kernel toggling takes micro-seconds, restart takes
{28\,s} on Llama-3.1-8B-Instruct and {60\,s} on larger models like Qwen3-32B, about 70\% of which is CUDA graph capture, as already observed by prior work \cite{foundry}. Throughout the restart the service is stopped: no requests are served and in-flight requests lose their
KV caches. Using eager mode also incurs prohibitively high overheads as we discuss in \cref{SEC:eval:cost}.

\begin{figure}[t]
\centering
\includegraphics[width=\columnwidth]
  {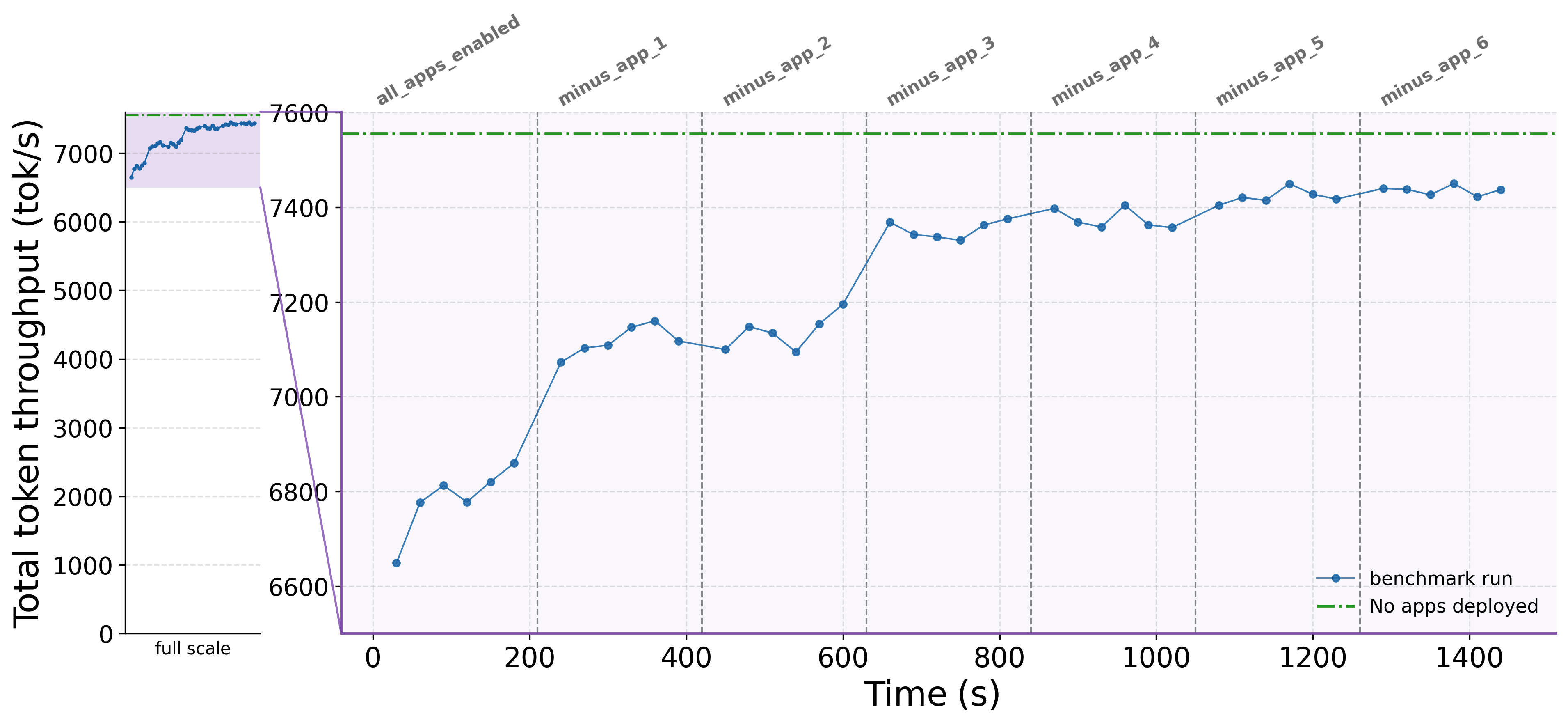}
\caption{Serving throughput while dynamically toggling co-deployed applications on
{Llama-3.1-8B-Instruct}. All six apps are initially active, and then 
deactivated one by one every  200 seconds (dashed lines) while the server is continuously
serving queries. The throughput grows by 9\% 
(6815 vs.\ 7435
~tok/s) than with all deactivated. 
}
\label{fig:eval:toggle-overhead}
\end{figure}

\para{Co-deployment of multiple applications}
\sys enables dynamic toggling of applications without service interruption. \cref{fig:eval:toggle-overhead} shows the serving throughput (\TTT) while serving requests and progressively deactivating six active applications in one instance. With all six \MI applications active, \TTT is 9.7\% below vanilla vLLM. Each deactivation reclaims its overhead share. Once all six are deactivated, performance matches vanilla vLLM.

This experiment shows that deactivated \MIapps \emph{impose no overheads}. This is by design: their nodes are
disabled in the captured graph so their kernels are not launched (\cref{SEC:runtime}). Operators can therefore keep multiple
\MIapps predeployed without performance cost, and toggle them on demand when needed \emph{without service interruption}. 

We also validated that toggling works correctly when different \MIapps are selected within the same batch.

\subsection{\acp{SAE}}
\label{SEC:eval:sae}

\begin{figure}[t]
  \centering
  \includegraphics[width=\columnwidth]
    {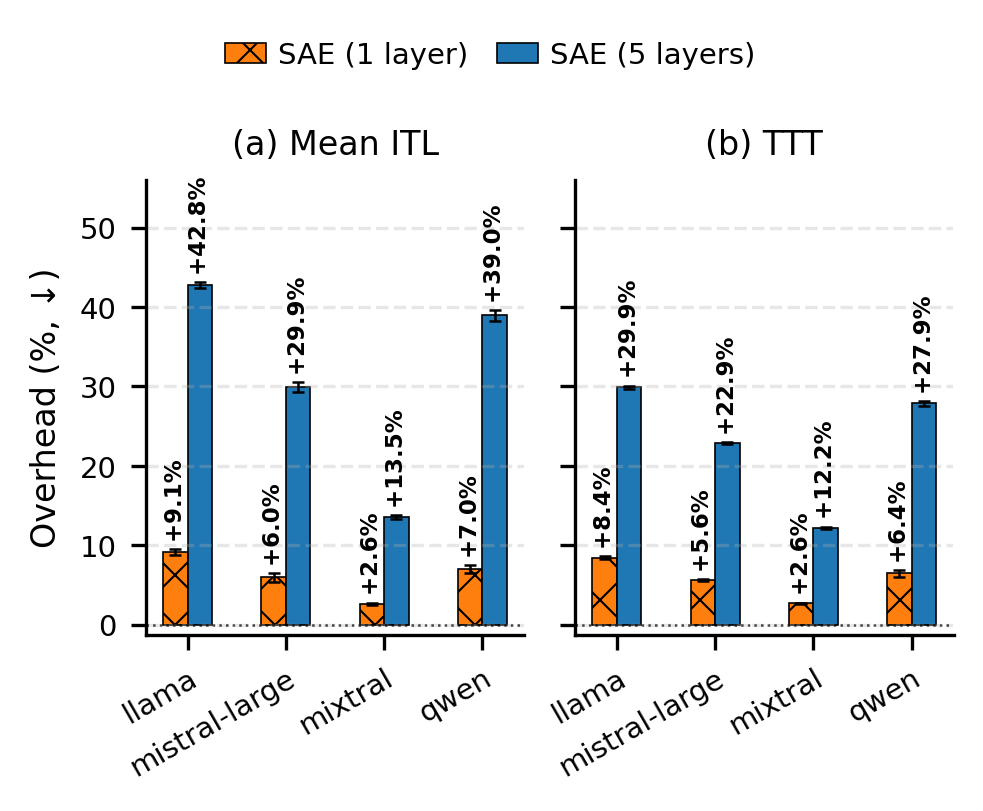}
    \vspace{-20pt}
  \caption{Overheads of running \ac{SAE} on one and five
  layers across different models: mean \ac{ITL} overhead and \ac{TTT} loss relative to vanilla. The setup is the same as in \cref{fig:eval:apps-ttt}.
  }
  \label{fig:eval:sae-cost}
\end{figure}

Unlike lightweight \MIapps discussed earlier, \acp{SAE} computations add significant computing and \emph{memory} load onto the system.   Each \ac{SAE} encodes the residual stream into a dictionary many times wider than the hidden
dimension, so it adds large matrix multiplications to every token
and potentially \emph{gigabytes of parameters} to every GPU. 
\cref{fig:eval:sae-cost} shows the performance for \acp{SAE} \MIapps across four models.  An
\ac{SAE} on a single layer increases mean \ITL by 2.6-9.1\%, and on
five layers by 13.5-42.8\%, reducing \TTT by up to 29.9\%. 
The performance is affected by both additional computations and \emph{the reduced number of in-flight requests}. This is because the memory needed for \acp{SAE} comes at the expense of KV cache, and smaller KV cache limits the number of active requests. 

\begin{figure}[t]
\centering
\includegraphics[width=\columnwidth,height=0.45\textheight,keepaspectratio]
  {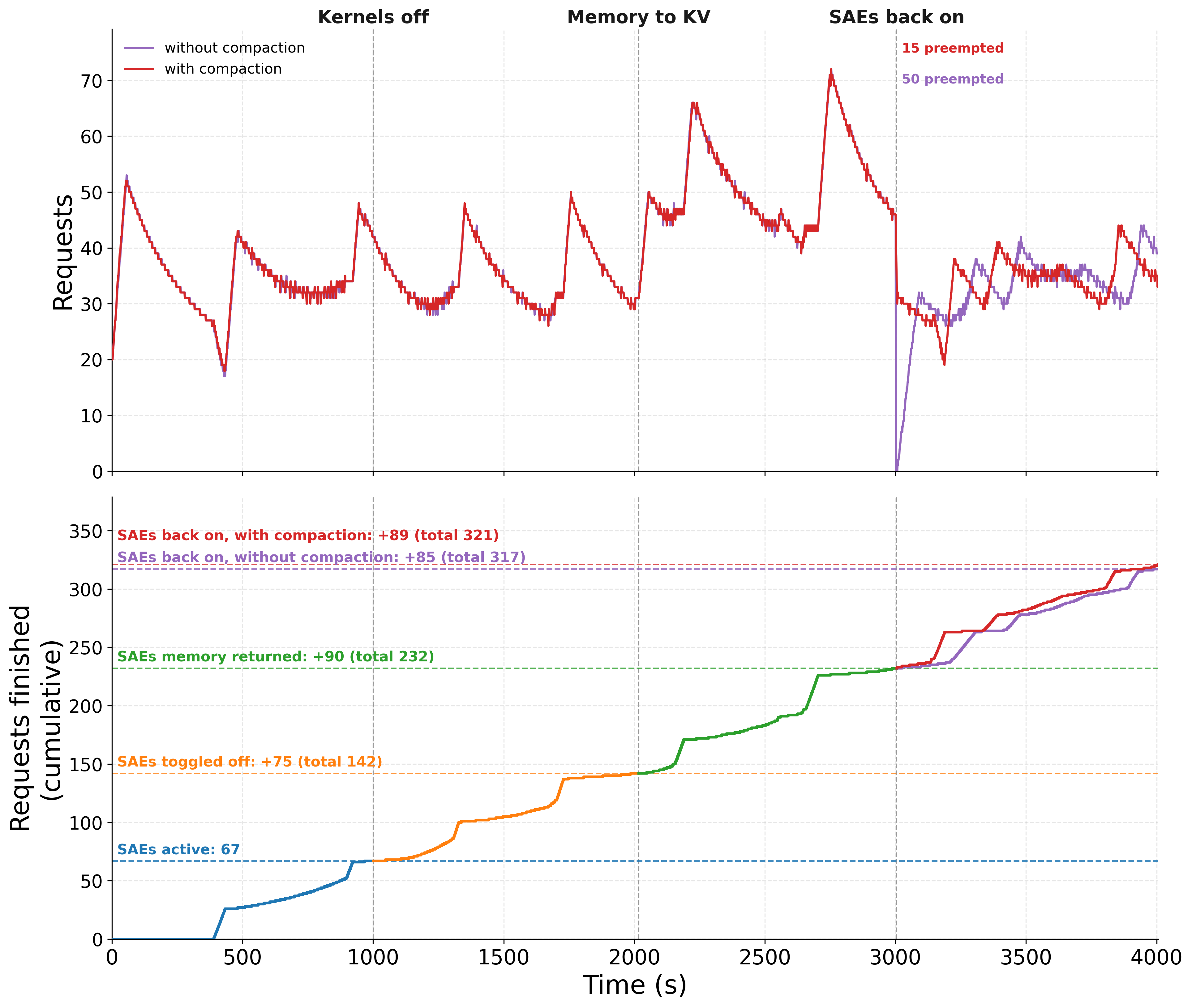}
\caption{Number of incomplete (top) and completed (bottom) requests while toggling \ac{SAE} \MIapps, with and without memory compaction and younger request preemption.%At t=1000 the \ac{SAE} kernels are toggled off but no memory is reclaimed, and at t=2000 their buffers are reclaimed by KV cache. At t=3000 the \acp{SAE} are reactivated and their memory is taken back from the KV cache, with and without KV cache compaction (inset).
}
\label{fig:eval:reclaim-throughput}
\end{figure}

To allow \acp{SAE} deployment in production, \sys offers the option to toggle them on demand, while also reclaiming the memory of inactive \acp{SAE} back to the KV cache. 
\cref{fig:eval:reclaim-throughput} shows the number of requests served (top) and 
completed (bottom).
with five \ac{SAE} applications deployed on
Llama-3.1-70B-Instruct running on four A100-80GB GPUs. \acp{SAE} occupy {11.5\,GB} on each GPU. 
Input/output length is fixed at {6k/6k} tokens to stress the KV cache.

We then deactivate SAEs in two stages: first we only disable their kernels (at $t$=1000) without reclaiming memory, and then reclaim memory and replenish the KV cache (at $t$=2000). Finally, we reactivate the SAEs (at $t$=3000) and reallocate their memory, reducing the KV cache again.
 
We observe that the request completion rate improves by 12\% when kernels are disabled and then by additional 20\% when memory gets reclaimed (completing 90 requests).
Further analysis shows that
returning memory to the KV
cache grows it from {20K} to {30K} blocks, and the engine admits
reaching about 50 concurrent requests vs. 35 when all SAEs are active.
When the SAEs are reactivated, \sys shrinks the KV cache back. 
Note that the number of running requests falls back to 
the same level as before deactivation, and the engine keeps serving
requests throughout.
We observe that our preemption policy and compaction is effective: activating SAEs preempts only 15 vs. \emph{all} running requests for the naive approach.

\subsection{State-of-the-art alternatives} \label{SS:eval:OtherComp}
\label{SEC:eval:cost}

%-----------------------------------------------------------------

\begin{figure}[t]
  \centering
  \begin{subfigure}[t]{\columnwidth}
    \centering
    \includegraphics[width=\columnwidth]
      {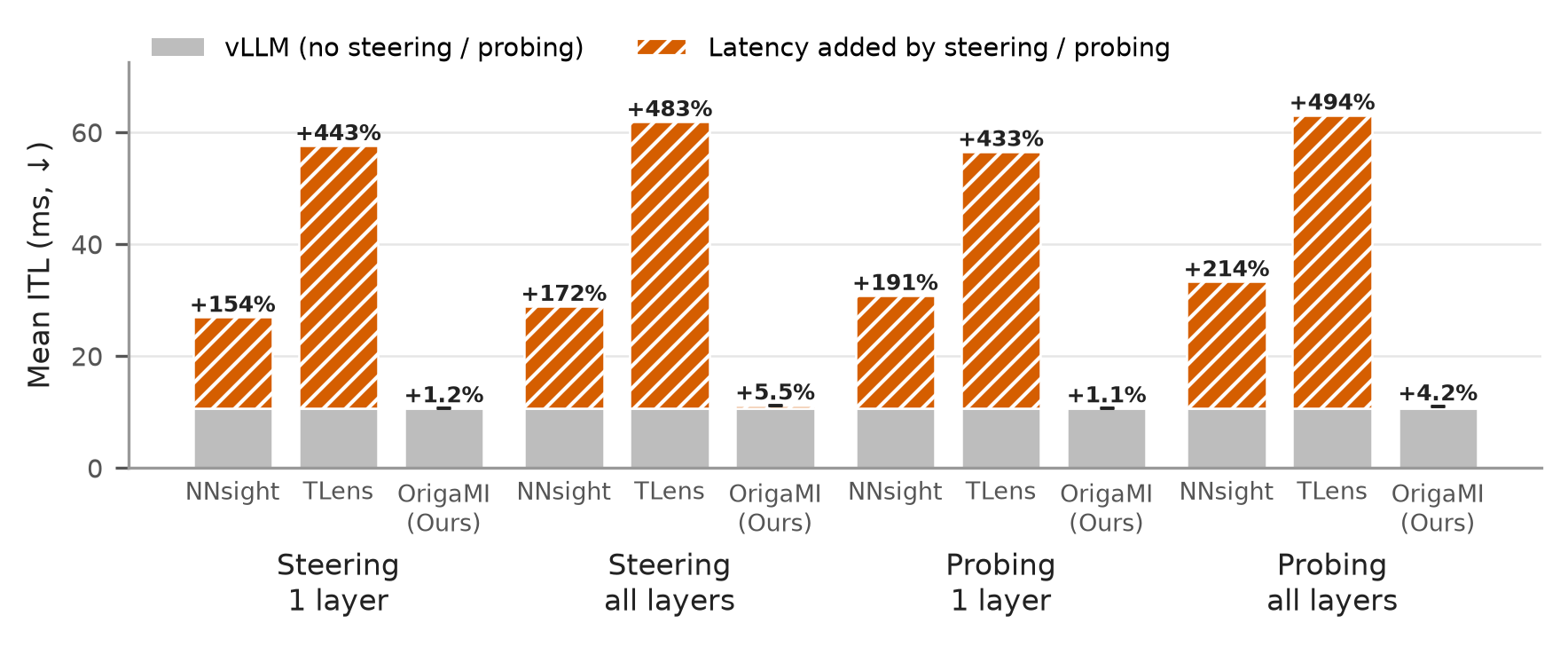}
    \caption{Mean \acs{ITL}: \NNSIGHT, \TLENS }
    \label{fig:eval:frameworks-itl}
  \end{subfigure}
  %\vspace{-0.5em}
  \begin{subfigure}[t]{\columnwidth}
    \centering
    \includegraphics[width=\columnwidth]
      {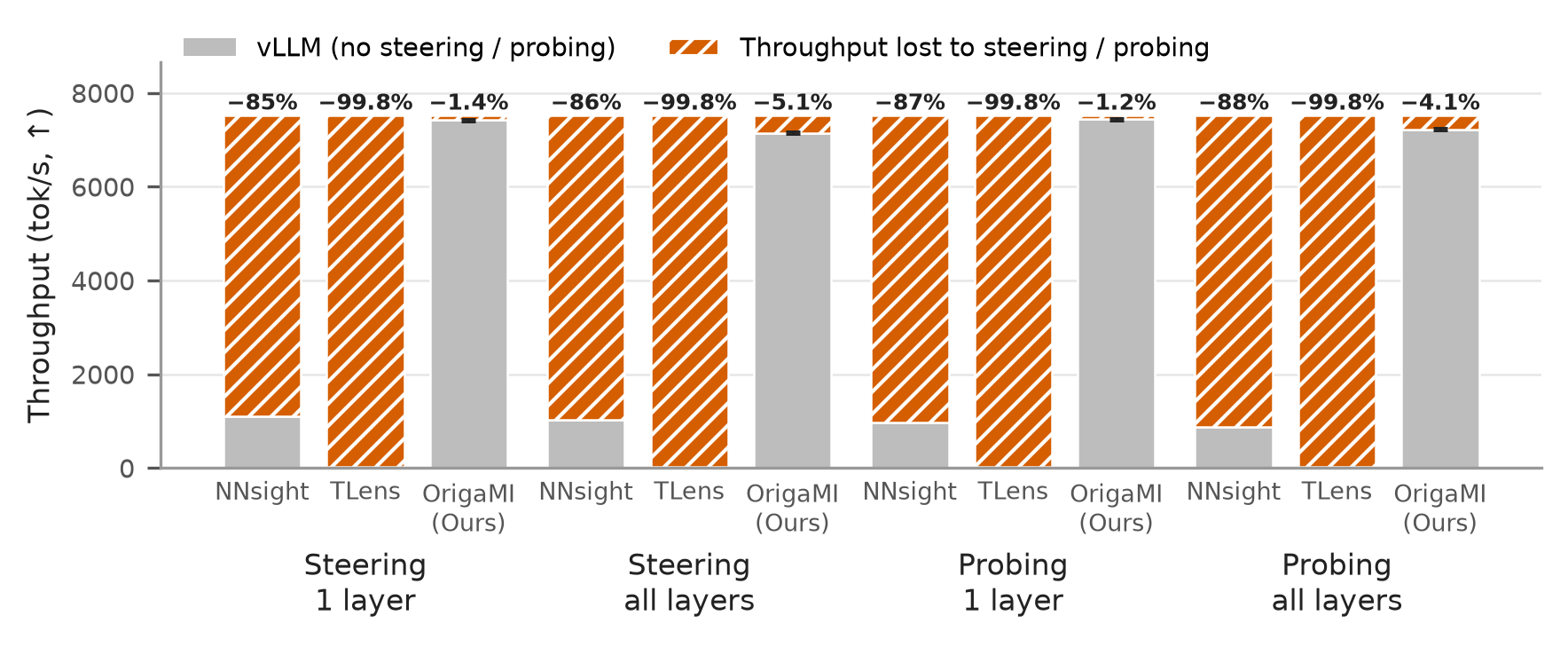}
    \caption{\acs{TTT}: \NNSIGHT, \TLENS}
    \label{fig:eval:frameworks-ttt}
  \end{subfigure}
  \begin{subfigure}[t]{\columnwidth}
  \centering
  \includegraphics[width=\columnwidth]
    {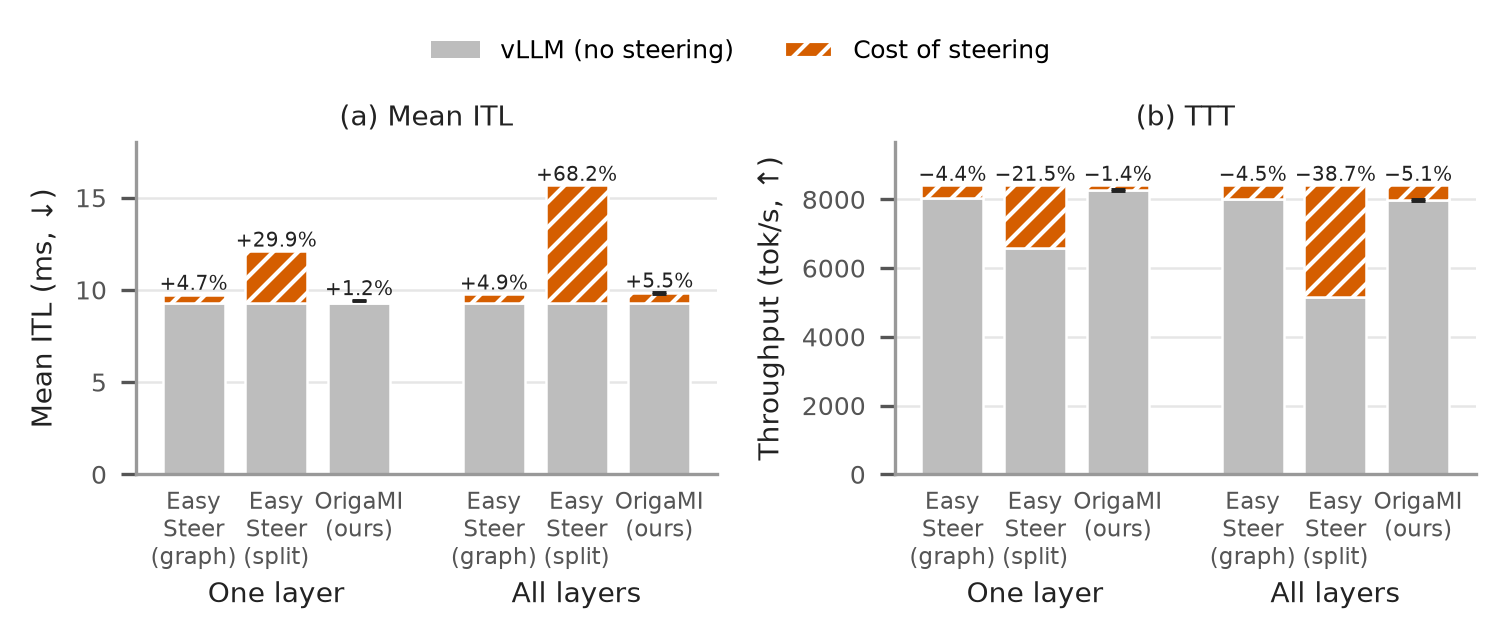}
  \caption{\ES with CUDA graphs}
  \label{fig:eval:easysteer-graphs}
\end{subfigure}
  \vspace{-6pt}
  \caption{Steering in Llama-3.1-8B-Instruct
  on \ES with CUDA graphs, \NNSIGHT, \TLENS~and \sys. Gray: \VLLM without steering; hatched:
  the overhead of a framework.}
  \label{fig:eval:frameworks}
\end{figure}

We distinguish between frameworks that always employ eager execution (\NNSIGHT~\cite{fiottokaufman2024nnsight} and \TLENS~\cite{TransformerLens}), general \ES~\cite{EasySteer} and those that use CUDA graphs (\ES-CUDA graphs~\cite{EasySteerCudaGraphs}).
We note that none of the existing solutions supports \MIapp toggling and memory management, so \sys provides a  superset of their \MIapp management capabilities. 

\para{Eager frameworks} 
%Most existing MI tooling was built for rapid research iteration, not for production inference.
\cref{fig:eval:frameworks}(a,b) shows the overheads of \NNSIGHT, \TLENS~and \sys when performing steering (write primitive) or deploying a probe (read primitive), over a single layer and all layers. These results are indicative of the expected overheads incurred by any \MI application using the same primitives. 

We observe that these frameworks suffer from a prohibitively high slowdown for both primitives. For example,  mean \ac{ITL} is affected by 154\% and 443\% for steering a single layer for \NNSIGHT~and \TLENS~ respectively. These overheads largely preclude the use of \MIapps in production. Instead, \sys incurs only 1.2\% slowdown, more than two orders of magnitude lower than these alternatives, and makes \MIapps production deployment feasible. Similar results are observed for general \ES (see Appendix,\cref{fig:appx:easysteer-apps}).

The overheads stem from the eager execution approach. The
eager frameworks introduce the host code for implementing \MIapps, breaking the CUDA graph optimization and resorting to launching all GPU kernels individually from the CPU. \sys, on the other hand, allows \MIapp implementation that is compatible with CUDA graphs.

\para{\ES} 
\label{SS:eval:ESComp}
\ES-CUDA graphs offers a subset of the general \ES steering functions that can run via CUDA graphs~\citep{EasySteerCudaGraphs}.
However, it has a number of limitations. 
The optimized mode has strict constraints and cannot be used for some \MIapps supported by \sys, such as probes \cite{orgad2025llmsknowshowintrinsic}, conditional steering \cite{programmingrefusalconditionalactivation}, and MoE router logits steering \cite{fayyaz2026steermoe}.

We chose a simple steering application supported by the \ES CUDA graph mode, and compare its performance with the same logic implemented in \sys. For both single-layer and all-layer steering, the two systems incur
comparable low overheads (\cref{fig:eval:easysteer-graphs}).
Note that \ES runs simple single-vector steering inside the model's full CUDA graph (in\_graph),
but more complex interventions fall back to a \texttt{split} mode, where
graph-splitting operator runs eagerly from Python between piecewise CUDA
graph segments, reducing throughput by up to 39\%.

We conclude that \sys achieves equivalent performance to \ES where \ES is compatible. 
Unlike \ES, however, it retains high performance for a much wider range of \MIapps and adds multi-app flexibility \ES lacks.

\ignore{
\begin{table}[t]
\centering
\tabletextsize
\begin{tabular}{lcc}
\toprule
Capability & \ES (CUDA graphs) & \sys \\
\midrule
Write                                    & \checkmark & \checkmark \\
Conditional Write on the GPU             & \tbd{?}    & \checkmark \\
Several applications in one instance     & \tbd{?}    & \checkmark \\
Per-request selection in a mixed batch   & \tbd{?}    & \checkmark \\
Activation/deactivation without restart  & \tbd{?}    & \checkmark \\
Memory reclamation to the KV cache       & \tbd{?}    & \checkmark \\
Attention scores with \ac{FA}            & \tbd{?}    & \checkmark \\
Tensor-parallel Reads                    & \tbd{?}    & \checkmark \\
\bottomrule
\end{tabular}
\caption{Capabilities of \ES with CUDA graphs~\citep{EasySteerCudaGraphs}
and of \sys.}
\label{tab:eval:capabilities}
\end{table}
}
%-------------------------------------------------------------

\section{Related Work}
\label{SEC:related}

\para{Existing MI serving Tooling}
\TLENS{} and \NNSIGHT{} expose model internals for flexible research
workflows~\cite{nanda2022transformerlens,fiottokaufman2024nnsight}.
Their host-side execution path incurs substantially higher inter-token
latency than optimized \VLLM serving in our comparison
(\cref{fig:eval:frameworks}). \ES integrates steering with
\VLLM~\cite{xu2025easysteer}; with CUDA graphs, its measured steering
overhead is comparable to \sys on a limited amount of scenarios (\cref{fig:eval:easysteer-graphs}). Its
interface, however, specializes in steering rather than a managed
portfolio of heterogeneous read and write apps. Moreover, it does not allow a dynamic deployment of multiple apps in parallel.

\para{Adapter dynamic serving}
Multi-adapter systems such as Punica and S-LoRA share a base model
across requests while selecting LoRA updates per
request~\cite{Punica,S-Lora}. However LoRA adapter differ by data, not logic. They optimize a common update form at known
layers. Which is handled by a dedicated selection kernel. Whereas the apps in \cref{tab:MIAPPSTab} additionally differ in their
primitives, triggers, and submodule loci, so adapter selection alone
does not express their execution requirements.

\ignore{
\para{KV cache memory management} vAttention~\cite{vattention}
manages the KV cache through CUDA virtual memory management, mapping
physical memory beneath reserved virtual addresses. It does so to keep
each request's KV cache virtually contiguous, so attention kernels
need no paging support. \sys uses the same mechanism for a different
purpose: moving physical memory between applications and the KV cache
beneath the fixed addresses that captured CUDA graphs hold, upon app deactivation or reactivation.
}
\section{Limitations and future work}
\label{SEC:limitations}

Our prototype remains limited in several aspects which we leave for future work.

\para{Model and framework portability}
First, extending \sys to new model architectures currently requires manual hook placement, yet in our experience this is relatively straightforward.
Further, we prototype \sys on \VLLM, but our preliminary analysis indicates that porting to SGLang and Ollama should not pose fundamental design challenges.

\para{MI app isolation}
\sys does not enforce strict memory isolation between concurrent MI applications, and custom user-provided kernels are not statically verified for CUDA-graph safety. These features would be necessary for production systems.

\para{Runtime and performance}
\sys currently does not support invocation of different \MIapps for requests with the same prefix cache. In addition, some \MIapps targeting intermediate activations inevitably break certain operator fusions, incurring a localized performance penalty. This can be mitigated by replacing the split kernels with fused ones when an \MIapp is inactive, but our current implementation does not support that. \sys memory management can be further improved to reduce memory reallocation costs.
Within the attention backend, \FI extracts attention scores based on thresholds; extending this to target specific token positions does not require structural changes to the kernel, however.
\section{Conclusion}
\label{SEC:conclusion}

This paper presents \sys, a serving-native platform that introduces an \MIapp abstraction and makes it the
first-class citizen in LLM inference engines. By analyzing state-of-the-art \MI applications, we formulated an execution abstraction that
formalizes their interface with the model. \sys
introduces a compiler that transforms the declarative \DSL definitions into
optimized, CUDA graph-compatible execution paths with a runtime that can dynamically
toggle \MIapps while managing GPU memory under variable workloads. We introduced
FlashInspect kernel that exposes runtime attention scores
without sacrificing fused-kernel throughput. These contributions bridge the
gap between mechanistic interpretability and production-grade serving, making
continuous, in-flight model introspection and runtime steering practical at
scale.

\bibliography{Bibs/Codex,Bibs/references,Bibs/manual}

@misc{nanda2022transformerlens,
  title = {TransformerLens},
  author = {Nanda, Neel and Bloom, Joseph},
  year = {2022},
  howpublished = {\url{https://github.com/TransformerLensOrg/TransformerLens}},
  note = {Software library citation from the project README; no archival paper located.}
}

@inproceedings{fiottokaufman2024nnsight,
  title={NNsight and NDIF: Democratizing Access to Open-Weight Foundation
    Model Internals},
  author={Fiotto-Kaufman, Jaden and Loftus, Alexander and Todd, Eric and
    Brinkmann, Jannik and Pal, Koyena and Troitskii, Dmitrii and Ripa,
    Michael and Belfki, Adam and Rager, Can and Juang, Caden and Mueller,
    Aaron and Marks, Samuel and Sharma, Arnab Sen and Lucchetti, Francesca
    and Prakash, Nikhil and Brodley, Carla and Guha, Arjun and Bell,
    Jonathan and Wallace, Byron and Bau, David},
  booktitle={International Conference on Learning Representations},
  year={2025},
  url={https://proceedings.iclr.cc/paper_files/paper/2025/hash/e6c65eb9b56719c1aa45ff73874de317-Abstract-Conference.html}
}

@inproceedings{zheng2024sglangfull,
  title={SGLang: Efficient Execution of Structured Language Model Programs},
  author={Zheng, Lianmin and Yin, Liangsheng and Xie, Zhiqiang and Sun,
    Chuyue and Huang, Jeff and Yu, Cody Hao and Cao, Shiyi and Kozyrakis,
    Christos and Stoica, Ion and Gonzalez, Joseph E. and Barrett, Clark
    and Sheng, Ying},
  booktitle={Advances in Neural Information Processing Systems},
  volume={37},
  pages={62557--62583},
  year={2024},
  url={https://proceedings.nips.cc/paper_files/paper/2024/hash/724be4472168f31ba1c9ac630f15dec8-Abstract-Conference.html}
}

@inproceedings{xu2025easysteer,
  title = {EasySteer: A Unified Framework for High-Performance and Extensible
    {LLM} Steering},
  author = {Xu, Haolei and Mei, Xinyu and Yan, Yuchen and Zhou, Rui and Zhang,
    Wenqi and Lu, Weiming and Zhuang, Yueting and Shen, Yongliang},
  booktitle = {Proceedings of the 2026 Conference on Empirical Methods in
    Natural Language Processing: System Demonstrations},
  year = {2026},
  pages = {24--35},
  address = {Budapest, Hungary},
  publisher = {Association for Computational Linguistics},
  url = {https://aclanthology.org/2026.emnlp-demos.3/}
}

@article{bentov2025universal,
  title = {Universal Jailbreak Suffixes Are Strong Attention Hijackers},
  author = {Ben-Tov, Matan and Geva, Mor and Sharif, Mahmood},
  journal = {arXiv preprint arXiv:2506.12880},
  year = {2025},
  note = {arXiv preprint used because no published venue was found.}
}

@inproceedings{wang2025sadi,
  title = {Semantics-Adaptive Activation Intervention for LLMs via Dynamic
    Steering Vectors},
  author = {Wang, Weixuan and Yang, Jingyuan and Peng, Wei},
  booktitle = {The Thirteenth International Conference on Learning
    Representations},
  year = {2025},
  url = {https://openreview.net/forum?id=8WQ7VTfPTl}
}

@inproceedings{hojer2025representation,
  title = {Improving Reasoning Performance in Large Language Models via
    Representation Engineering},
  author = {H{\o}jer, Bertram and Jarvis, Oliver Simon and Heinrich, Stefan},
  booktitle = {The Thirteenth International Conference on Learning
    Representations},
  year = {2025},
  url = {https://openreview.net/forum?id=IssPhpUsKt}
}

@inproceedings{ferrando2025entity,
  title = {Do I Know This Entity? Knowledge Awareness and Hallucinations in
    Language Models},
  author = {Ferrando, Javier and Obeso, Oscar Balcells and Rajamanoharan,
    Senthooran and Nanda, Neel},
  booktitle = {The Thirteenth International Conference on Learning
    Representations},
  year = {2025},
  url = {https://openreview.net/forum?id=WCRQFlji2q}
}

@inproceedings{cheng2025backtracking,
  title = {Steering When Necessary: Flexible Steering Large Language Models
    with Backtracking},
  author = {Cheng, Zifeng and Gan, Jinwei and Jiang, Zhiwei and Wang, Cong and
    Yin, Yafeng and Luo, Xiang and Fu, Yuchen and Gu, Qing},
  booktitle = {Advances in Neural Information Processing Systems},
  year = {2025},
  url = {https://openreview.net/forum?id=l75RyRcevf}
}

@inproceedings{fayyaz2026steermoe,
  title = {Steering {MoE} {LLM}s via Expert ({De})Activation},
  author = {Fayyaz, Mohsen and Modarressi, Ali and Deilamsalehy, Hanieh and
    Dernoncourt, Franck and Rossi, Ryan A. and Bui, Trung and Schuetze,
    Hinrich and Peng, Nanyun},
  booktitle = {The Fourteenth International Conference on Learning
    Representations},
  year = {2026},
  url = {https://openreview.net/forum?id=v5Yl9V8rJs}
}

@inproceedings{wang2025twoexperts,
  title = {Two Experts Are All You Need for Steering Thinking: Reinforcing
    Cognitive Effort in {MoE} Reasoning Models Without Additional Training},
  author = {Wang, Mengru and Chen, Xingyu and Wang, Yue and He, Zhiwei and Xu,
    Jiahao and Liang, Tian and Liu, Qiuzhi and Yao, Yunzhi and Wang, Wenxuan
    and Ma, Ruotian and Mi, Haitao and Zhang, Ningyu and Tu, Zhaopeng and Li,
    Xiaolong and Yu, Dong},
  booktitle = {Advances in Neural Information Processing Systems},
  year = {2025},
  url = {https://openreview.net/forum?id=x7fCiuCCAu}
}

@misc{anthropic2025sonnetcard,
  author = {{Anthropic}},
  title = {{Claude Sonnet 4.5 System Card}},
  year = {2025},
  month = sep,
  howpublished = {\href{https://www-cdn.anthropic.com/963373e433e489a87a10c823c52a0a013e9172dd.pdf}{Anthropic PDF}},
  note = {Primary model system card, Section 7.6, on pre-release white-box
    interpretability investigations.}
}

@misc{anthropic2026mythoscard,
  author = {{Anthropic}},
  title = {{Claude Mythos Preview System Card}},
  year = {2026},
  month = apr,
  howpublished = {\href{https://www-cdn.anthropic.com/08ab9158070959f88f296514c21b7facce6f52bc.pdf}{Anthropic PDF}},
  note = {Primary model system card, Section 4.5, on white-box analyses of
    model internals.}
}

@misc{goodfireRakutenSAE2025,
  title = {Deploying Interpretability to Production with {Rakuten}: {SAE}
    Probes for {PII} Detection},
  author = {Nguyen, Nam and Deng, Myra and Gala, Dhruvil and Byun, Michael
    and Naruse, Kenta and Virgo, Felix Giovanni and Hazra, Dron and Gorton,
    Liv and Balsam, Daniel and McGrath, Thomas and Takei, Mio and Kaji,
    Yusuke},
  year = {2025},
  month = oct,
  howpublished = {Goodfire Research},
  url = {https://www.goodfire.com/research/rakuten-sae-probes-for-pii-detection},
  note = {Industry research report on deployed SAE probes using a sidecar
    model for Rakuten's PII detection.}
}

@misc{kramarGeminiProbes2026,
  title = {Building Production-Ready Probes For {Gemini}},
  author = {Kram{\'a}r, J{\'a}nos and Engels, Joshua and Wang, Zheng and
    Chughtai, Bilal and Shah, Rohin and Nanda, Neel and Conmy, Arthur},
  year = {2026},
  eprint = {2601.11516},
  archivePrefix = {arXiv},
  url = {https://arxiv.org/abs/2601.11516},
  note = {arXiv preprint cited for Google DeepMind's own report of probes
    deployed in user-facing Gemini and production distribution shifts;
    industry research exception.}
}

@inproceedings{cunninghamConstitutionalClassifiers2026,
  title = {Constitutional Classifiers++: Efficient Production-Grade Defenses
    against Universal Jailbreaks},
  author = {Cunningham, Hoagy and Wei, Jerry and Wang, Zihan and Persic,
    Andrew and Peng, Alwin and Abderrachid, Jordan and Agarwal, Raj and
    Chen, Bobby and Dau, Andy and Dimitriev, Alek and Howard, Logan and
    Hua, Yijin and Gilson, Rob and Lin, Mu and Liu, Christopher and
    Mikulik, Vladimir and Mittapalli, Rohit and O'Hara, Clare and Pan,
    Jin and Saxena, Nikhil and Silverstein, Alex and Song, Yue and Yu,
    Xunjie and Zhou, Giulio and Leike, Jan and Kaplan, Jared and Perez,
    Ethan and Sharma, Mrinank},
  booktitle = {The Fourteenth International Conference on Learning
    Representations},
  year = {2026},
  url = {https://proceedings.iclr.cc/paper_files/paper/2026/hash/4f697305ef1f868ad77c3c0027989a6f-Abstract-Conference.html}
}

@inproceedings{scialanga-etal-2025-sake,
    title = "{SAKE}: Steering Activations for Knowledge Editing",
    author = "Scialanga, Marco  and
      Laugel, Thibault  and
      Grari, Vincent  and
      Detyniecki, Marcin",
    editor = "Che, Wanxiang  and
      Nabende, Joyce  and
      Shutova, Ekaterina  and
      Pilehvar, Mohammad Taher",
    booktitle = "Proceedings of the 63rd Annual Meeting of the Association for Computational Linguistics (Volume 1: Long Papers)",
    month = jul,
    year = "2025",
    address = "Vienna, Austria",
    publisher = "Association for Computational Linguistics",
    url = "https://aclanthology.org/2025.acl-long.777/",
    doi = "10.18653/v1/2025.acl-long.777",
    pages = "15966--15978",
    ISBN = "979-8-89176-251-0"
}

@article{flashattention,
  title={Flashattention: Fast and memory-efficient exact attention with io-awareness},
  author={Dao, Tri and Fu, Dan and Ermon, Stefano and Rudra, Atri and R{\'e}, Christopher},
  journal={Advances in neural information processing systems},
  volume={35},
  pages={16344--16359},
  year={2022}
}

@inproceedings{
orgad2025llmsknowshowintrinsic,
title={{LLM}s Know More Than They Show: On the Intrinsic Representation of {LLM} Hallucinations},
author={Hadas Orgad and Michael Toker and Zorik Gekhman and Roi Reichart and Idan Szpektor and Hadas Kotek and Yonatan Belinkov},
booktitle={The Thirteenth International Conference on Learning Representations},
year={2025},
url={https://openreview.net/forum?id=KRnsX5Em3W}
}

@inproceedings{PagedAttention,
  title={Efficient memory management for large language model serving with pagedattention},
  author={Kwon, Woosuk and Li, Zhuohan and Zhuang, Siyuan and Sheng, Ying and Zheng, Lianmin and Yu, Cody Hao and Gonzalez, Joseph and Zhang, Hao and Stoica, Ion},
  booktitle={Proceedings of the 29th symposium on operating systems principles},
  pages={611--626},
  year={2023}
}

@inproceedings {Orca,
author = {Gyeong-In Yu and Joo Seong Jeong and Geon-Woo Kim and Soojeong Kim and Byung-Gon Chun},
title = {Orca: A Distributed Serving System for {Transformer-Based} Generative Models},
booktitle = {16th USENIX Symposium on Operating Systems Design and Implementation (OSDI 22)},
year = {2022},
isbn = {978-1-939133-28-1},
address = {Carlsbad, CA},
pages = {521--538},
url = {https://www.usenix.org/conference/osdi22/presentation/yu},
publisher = {USENIX Association},
month = jul
}

@misc{TransformerLens,
      title={TransformerLens}, 
      author={Neel Nanda},
      year={2023},
      url={https://transformerlensorg.github.io/TransformerLens/},
      note = {Accessed: 2026-02-14}
}

@misc{LlamaScope,
      title={Llama Scope: Extracting Millions of Features from Llama-3.1-8B with Sparse Autoencoders}, 
      author={Zhengfu He and Wentao Shu and Xuyang Ge and Lingjie Chen and Junxuan Wang and Yunhua Zhou and Frances Liu and Qipeng Guo and Xuanjing Huang and Zuxuan Wu and Yu-Gang Jiang and Xipeng Qiu},
      year={2024},
      eprint={2410.20526},
      archivePrefix={arXiv},
      primaryClass={cs.LG},
      url={https://arxiv.org/abs/2410.20526}, 
}

@inproceedings{
programmingrefusalconditionalactivation,
title={Programming Refusal with Conditional Activation Steering},
author={Bruce W. Lee and Inkit Padhi and Karthikeyan Natesan Ramamurthy and Erik Miehling and Pierre Dognin and Manish Nagireddy and Amit Dhurandhar},
booktitle={The Thirteenth International Conference on Learning Representations},
year={2025},
url={https://openreview.net/forum?id=Oi47wc10sm}
}

@article{arditi2024refusallanguagemodelsmediated,
  title={Refusal in language models is mediated by a single direction},
  author={Arditi, Andy and Obeso, Oscar and Syed, Aaquib and Paleka, Daniel and Panickssery, Nina and Gurnee, Wes and Nanda, Neel},
  journal={Advances in Neural Information Processing Systems},
  volume={37},
  pages={136037--136083},
  year={2024}
}

@inproceedings{
chen2025sealsteerablereasoningcalibration,
title={{SEAL}: Steerable Reasoning Calibration of Large Language Models for Free},
author={Runjin Chen and Zhenyu Zhang and Junyuan Hong and Souvik Kundu and Zhangyang Wang},
booktitle={Second Conference on Language Modeling},
year={2025},
url={https://openreview.net/forum?id=klPszYDIRT}
}

@article{li2023inference,
  title={Inference-time intervention: Eliciting truthful answers from a language model},
  author={Li, Kenneth and Patel, Oam and Vi{\'e}gas, Fernanda and Pfister, Hanspeter and Wattenberg, Martin},
  journal={Advances in Neural Information Processing Systems},
  volume={36},
  pages={41451--41530},
  year={2023}
}

@inproceedings{
soo2025interpretable,
title={Interpretable Steering of Large Language Models with Feature Guided Activation Additions},
author={Samuel Soo and Wesley Teng and Chandrasekaran Balaganesh and Tan Guoxian and Ming YAN},
booktitle={ICLR 2025 Workshop on Building Trust in Language Models and Applications},
year={2025},
url={https://openreview.net/forum?id=swRxS7s4rB}
}

@inproceedings{
stickland2024steering,
title={Steering Without Side Effects: Improving Post-Deployment Control of Language Models},
author={Asa Cooper Stickland and Alexander Lyzhov and Jacob Pfau and Salsabila Mahdi and Samuel R. Bowman},
booktitle={Neurips Safe Generative AI Workshop 2024},
year={2024},
url={https://openreview.net/forum?id=tfXIZ8P4ZU}
}

@misc{
turner2025steering,
title={Steering Language Models with Activation Engineering},
author={Alexander Matt Turner and Lisa Thiergart and Gavin Leech and David Udell and Juan J Vazquez and Ulisse Mini and Monte MacDiarmid},
year={2025},
url={https://openreview.net/forum?id=2XBPdPIcFK}
}

@inproceedings{farooq2025sentiment,
  title={Sentiment steering in large language models via activation vector manipulation},
  author={Farooq, Misbah and De Silva, Varuna and Rahulamathavan, Rahul and Shi, Xiyu},
  booktitle={2025 25th International Conference on Digital Signal Processing (DSP)},
  pages={1--5},
  year={2025},
  organization={IEEE}
}

@misc{hu2024toxicitydetectionfree,
      title={Toxicity Detection for Free}, 
      author={Zhanhao Hu and Julien Piet and Geng Zhao and Jiantao Jiao and David Wagner},
      year={2024},
      eprint={2405.18822},
      archivePrefix={arXiv},
      primaryClass={cs.CL},
      url={https://arxiv.org/abs/2405.18822}, 
}

@inproceedings{
wang2025semanticsadaptive,
title={Semantics-Adaptive Activation Intervention for {LLM}s via Dynamic Steering Vectors},
author={Weixuan Wang and JINGYUAN YANG and Wei Peng},
booktitle={The Thirteenth International Conference on Learning Representations},
year={2025},
url={https://openreview.net/forum?id=8WQ7VTfPTl}
}

@inproceedings{
wang2026two,
title={Two Experts Are All You Need for Steering Thinking: Reinforcing Cognitive Effort in MoE Reasoning Models Without Additional Training},
author={Mengru Wang and Xingyu Chen and Yue Wang and Zhiwei He and Jiahao Xu and Tian Liang and Qiuzhi Liu and Yunzhi Yao and Wenxuan Wang and Ruotian Ma and Haitao Mi and Ningyu Zhang and Zhaopeng Tu and Xiaolong Li and Dong Yu},
booktitle={The Thirty-ninth Annual Conference on Neural Information Processing Systems},
year={2026},
url={https://openreview.net/forum?id=x7fCiuCCAu}
}

@inproceedings{rimsky-etal-2024-steering,
    title = "Steering Llama 2 via Contrastive Activation Addition",
    author = "Rimsky, Nina  and
      Gabrieli, Nick  and
      Schulz, Julian  and
      Tong, Meg  and
      Hubinger, Evan  and
      Turner, Alexander",
    editor = "Ku, Lun-Wei  and
      Martins, Andre  and
      Srikumar, Vivek",
    booktitle = "Proceedings of the 62nd Annual Meeting of the Association for Computational Linguistics (Volume 1: Long Papers)",
    month = aug,
    year = "2024",
    address = "Bangkok, Thailand",
    publisher = "Association for Computational Linguistics",
    url = "https://aclanthology.org/2024.acl-long.828/",
    doi = "10.18653/v1/2024.acl-long.828",
    pages = "15504--15522"
}

@inproceedings{wang-etal-2025-bridging,
    title = "Bridging the Language Gaps in Large Language Models with Inference-Time Cross-Lingual Intervention",
    author = "Wang, Weixuan  and
      Wu, Minghao  and
      Haddow, Barry  and
      Birch, Alexandra",
    editor = "Che, Wanxiang  and
      Nabende, Joyce  and
      Shutova, Ekaterina  and
      Pilehvar, Mohammad Taher",
    booktitle = "Proceedings of the 63rd Annual Meeting of the Association for Computational Linguistics (Volume 1: Long Papers)",
    month = jul,
    year = "2025",
    address = "Vienna, Austria",
    publisher = "Association for Computational Linguistics",
    url = "https://aclanthology.org/2025.acl-long.270/",
    doi = "10.18653/v1/2025.acl-long.270",
    pages = "5418--5433",
    ISBN = "979-8-89176-251-0"
}

@misc{vLLM-Bench-sweep-serve,
  author       = {{vLLM Project}},
  title        = {Benchmark Sweep: Serve},
  year         = {2026},
  howpublished  = {\url{https://docs.vllm.ai/en/latest/cli/bench/sweep/serve}},
  note         = {Accessed: 2026-06-08}
}

@misc{ShareGPT,
  author       = {{ShareGPT}},
  title        = {ShareGPT Vicuna Unfiltered Dataset},
  year         = {2026},
  howpublished  = {\url{https://huggingface.co/datasets/anon8231489123/ShareGPT_Vicuna_unfiltered}},
  note         = {Accessed: 2026-06-08}
}

@misc{EasySteerCudaGraphs,
  author       = {{EasySteer}},
  title        = {Server-level steering with CUDA graphs},
  year         = {2026},
  howpublished  = {\url{https://github.com/ZJU-REAL/EasySteer-vllm-v1/pull/3}},
  note         = {Accessed: 2026-06-11}
}

@inproceedings{
Charm,
title={Neural Message-Passing on Attention Graphs for Hallucination Detection},
author={Fabrizio Frasca and Guy Bar-Shalom and Yftah Ziser and Haggai Maron},
booktitle={The Fourteenth International Conference on Learning Representations},
year={2026},
url={https://openreview.net/forum?id=4twbqwV4br}
}

@inproceedings{Lookback_Lens,
    title = "Lookback Lens: Detecting and Mitigating Contextual Hallucinations in Large Language Models Using Only Attention Maps",
    author = "Chuang, Yung-Sung  and
      Qiu, Linlu  and
      Hsieh, Cheng-Yu  and
      Krishna, Ranjay  and
      Kim, Yoon  and
      Glass, James R.",
    editor = "Al-Onaizan, Yaser  and
      Bansal, Mohit  and
      Chen, Yun-Nung",
    booktitle = "Proceedings of the 2024 Conference on Empirical Methods in Natural Language Processing",
    month = nov,
    year = "2024",
    address = "Miami, Florida, USA",
    publisher = "Association for Computational Linguistics",
    url = "https://aclanthology.org/2024.emnlp-main.84/",
    doi = "10.18653/v1/2024.emnlp-main.84",
    pages = "1419--1436",
}

@inproceedings{
LLM-Check,
title={{LLM}-Check: Investigating Detection of Hallucinations in Large Language Models},
author={Gaurang Sriramanan and Siddhant Bharti and Vinu Sankar Sadasivan and Shoumik Saha and Priyatham Kattakinda and Soheil Feizi},
booktitle={The Thirty-eighth Annual Conference on Neural Information Processing Systems},
year={2024},
url={https://openreview.net/forum?id=LYx4w3CAgy}
}

@inproceedings{
FlashAttention-2,
title={FlashAttention-2: Faster Attention with Better Parallelism and Work Partitioning},
author={Tri Dao},
booktitle={The Twelfth International Conference on Learning Representations},
year={2024},
url={https://openreview.net/forum?id=mZn2Xyh9Ec}
}

@inproceedings{Punica,
 author = {Chen, Lequn and Ye, Zihao and Wu, Yongji and Zhuo, Danyang and Ceze, Luis and Krishnamurthy, Arvind},
 booktitle = {Proceedings of Machine Learning and Systems},
 editor = {P. Gibbons and G. Pekhimenko and C. De Sa},
 pages = {1--13},
 title = {Punica: Multi-Tenant LoRA Serving},
 url = {https://proceedings.mlsys.org/paper_files/paper/2024/file/054de805fcceb78a201f5e9d53c85908-Paper-Conference.pdf},
 volume = {6},
 year = {2024}
}

@inproceedings{S-Lora,
 author = {Sheng, Ying and Cao, Shiyi and Li, Dacheng and Hooper, Coleman and Lee, Nicholas and Yang, Shuo and Chou, Christopher and Zhu, Banghua and Zheng, Lianmin and Keutzer, Kurt and Gonzalez, Joseph E. and Stoica, Ion},
 booktitle = {Proceedings of Machine Learning and Systems},
 editor = {P. Gibbons and G. Pekhimenko and C. De Sa},
 pages = {296--311},
 title = {SLoRA: Scalable Serving of Thousands of LoRA Adapters},
 url = {https://proceedings.mlsys.org/paper_files/paper/2024/file/906419cd502575b617cc489a1a696a67-Paper-Conference.pdf},
 volume = {6},
 year = {2024}
}

@inproceedings{dlora,
  title={$\{$dLoRA$\}$: Dynamically orchestrating requests and adapters for $\{$LoRA$\}$$\{$LLM$\}$ serving},
  author={Wu, Bingyang and Zhu, Ruidong and Zhang, Zili and Sun, Peng and Liu, Xuanzhe and Jin, Xin},
  booktitle={18th USENIX Symposium on Operating Systems Design and Implementation (OSDI 24)},
  pages={911--927},
  year={2024}
}

@inproceedings{vattention,
  title={vattention: Dynamic memory management for serving llms without pagedattention},
  author={Prabhu, Ramya and Nayak, Ajay and Mohan, Jayashree and Ramjee, Ramachandran and Panwar, Ashish},
  booktitle={Proceedings of the 30th ACM International Conference on Architectural Support for Programming Languages and Operating Systems, Volume 1},
  pages={1133--1150},
  year={2025}
}

@article{foundry,
  title={Foundry: Template-Based CUDA Graph Context Materialization for Fast LLM Serving Cold Start},
  author={Liu, Xueshen and Wu, Yongji and Yao, Yuncheng and Zhuo, Danyang and Stoica, Ion and Mao, Z Morley},
  journal={arXiv preprint arXiv:2604.06664},
  year={2026}
}

@inproceedings{EasySteer,
    title = "{E}asy{S}teer: A Unified Framework for High-Performance and Extensible {LLM} Steering",
    author = "Xu, Haolei  and
      Mei, Xinyu  and
      Yan, Yuchen  and
      Zhou, Rui  and
      Zhang, Wenqi  and
      Lu, Weiming  and
      Zhuang, Yueting  and
      Shen, Yongliang",
    editor = "Agrawal, Sweta  and
      Abad, Alberto",
    booktitle = "Proceedings of the 2026 Conference on Empirical Methods in Natural Language Processing: System Demonstrations",
    month = oct,
    year = "2026",
    address = "Budapest, Hungary",
    publisher = "Association for Computational Linguistics",
    url = "https://aclanthology.org/2026.emnlp-demos.3/",
    pages = "24--35",
    ISBN = "979-8-89176-461-3",
}
\bibliographystyle{mlsys2025}

%\input{Sections/10_AI_usage_disc}

%%%%%%%%%%%%%%%%%%%%%%%%%%%%%%%%%%%%%%%%%%%%%%%%%%%%%%%%%%%%%%%%%%%%%%%%%%%%%%%
% SUPPLEMENTAL CONTENT AS APPENDIX AFTER REFERENCES
%%%%%%%%%%%%%%%%%%%%%%%%%%%%%%%%%%%%%%%%%%%%%%%%%%%%%%%%%%%%%%%%%%%%%%%%%%%%%%%
\appendix

\newpage

\section{Full Survey Mapping}
\label{appx:survey}

This appendix maps each surveyed \MIMethod to the
representative method listed in \cref{tab:MIAPPSTab}.

\begin{table*}[h]
\centering
\tabletextsize
\setlength{\tabcolsep}{6pt}
\renewcommand{\arraystretch}{1.05}
\begin{tabular}{
    p{0.02\textwidth}p{0.42\textwidth}p{0.16\textwidth}p{0.24\textwidth}}
\toprule
\# & Surveyed application (paper) & Representative app evaluated &
Notes / cluster rationale \\
\midrule
1 & {\textbf{Single Direction} (\citeauthor{arditi2024refusallanguagemodelsmediated}
  [\citeyear{arditi2024refusallanguagemodelsmediated}])} & {\textbf{App-1}} & Unconditional Multi-layer steering
 \\
2 & {Steering Language (\citeauthor{turner2025steering}
  [\citeyear{turner2025steering}])} & {App-1} &
Unconditional Multi-layer steering \\
3 & {Sentiment steering (\citeauthor{farooq2025sentiment}{}) [\citeyear{farooq2025sentiment}]} & {App-1} &
Unconditional Multi-layer steering \\
4 & {Bridging the Language Gaps in Large Language Models with Inference-Time Cross-Lingual Intervention (\citeauthor{wang-etal-2025-bridging}) [\citeyear{wang-etal-2025-bridging}]} & {App-1} &
Unconditional Multi-layer steering \\
5 & {\textbf{SEAL} (\citeauthor{chen2025sealsteerablereasoningcalibration}
[\citeyear{chen2025sealsteerablereasoningcalibration}])} & {\textbf{App-2}} & Token Conditional steering
 \\
6 & {\textbf{Conditional Refusal} (\citeauthor{programmingrefusalconditionalactivation}
  [\citeyear{programmingrefusalconditionalactivation}])} & {\textbf{App-3}} & Conditional steering
 \\
7 & {Semantics-Adaptive Activation Intervention for {LLM}s via Dynamic Steering Vectors (\citeauthor{wang2025semanticsadaptive}) [\citeyear{wang2025semanticsadaptive}]} & {App-3} &
Conditional steering \\
8 & {\textbf{Hallucination Probe} (\citeauthor{orgad2025llmsknowshowintrinsic}
  [\citeyear{orgad2025llmsknowshowintrinsic}])} & {\textbf{App-4}} & Logistic regression used on activations
 \\
9 & {Toxicity Detection for Free  (\citeauthor{hu2024toxicitydetectionfree}{}) [\citeyear{hu2024toxicitydetectionfree}]} & {App-4} &
Logistic regression used on activations \\
10 & {\textbf{SAKE} (\citeauthor{scialanga-etal-2025-sake}
  [\citeyear{scialanga-etal-2025-sake}])} & {\textbf{App-5}} &
 \\
11 & {Steering Llama 2 via Contrastive Activation Addition (\citeauthor{rimsky-etal-2024-steering}) [\citeyear{rimsky-etal-2024-steering}]} & {App-5} &
Unconditional write on a single layer \\
12 & {\textbf{Side-Effect-Free Steering} (\citeauthor{stickland2024steering}
  [\citeyear{stickland2024steering}])} & {\textbf{App-6}} &
Unconditional Multi-layer steering \\

13 & {\textbf{Truthful Intervention} (\citeauthor{li2023inference}
  [\citeyear{li2023inference}])} & {\textbf{App-7}} &
Unconditional Multi-layer steering inside attention submodule \\
14 & {\textbf{Universal prefixes} (\citeauthor{bentov2025universal}
  [\citeyear{bentov2025universal}])} & {\textbf{App-8}} &
Read at the last prompt token across multiple layers \\
15 & {Neural Message-Passing on Attention Graphs for Hallucination Detection (\citeauthor{Charm}) [\citeyear{Charm}]} & {App-8} &
Reads attention scores \\
16 & {Lookback Lens: Detecting and Mitigating Contextual Hallucinations in Large Language Models Using Only Attention Maps (\citeauthor{Lookback_Lens}) [\citeyear{Lookback_Lens}]} & {App-8} &
Reads attention scores \\
17 & {LLM-Check: Investigating Detection of Hallucinations in Large Language Models (\citeauthor{LLM-Check}) [\citeyear{LLM-Check}]} & {App-8} &
Reads attention scores \\
18 & {\textbf{SteerMoE} (\citeauthor{fayyaz2026steermoe}
  [\citeyear{fayyaz2026steermoe}])} & {\textbf{App-9}} &
MoE Router steering \\
19 & {Two Experts Are All You Need for Steering Thinking: Reinforcing Cognitive Effort in MoE Reasoning Models Without Additional Training (\citeauthor{wang2026two}) [\citeyear{wang2026two}]} & {App-9} &
MoE Router steering \\
20 & {\textbf{SAE} (\citeauthor{LlamaScope}
  [\citeyear{LlamaScope}])} & {\textbf{App-10}} &
Unconditional read on a single layer \\
\bottomrule
\end{tabular}
\caption{Surveyed-app to representative-app
mapping referenced in \cref{tab:MIAPPSTab}.}
\label{tab:survey_mapping}
\end{table*}

\section{Supported loci}
\label{appx:loci}

\cref{tab:loci} lists the submodules \sys exposes to \MIapps.

\begin{table*}[t]
\centering
\small
\begin{tabular}{l p{0.36\textwidth} l}
\toprule
Submodule name & Location in the forward pass & Example application \\
\midrule
{\texttt{post\_norm\_pre\_attn}} & {Attention block input} & {\-} \\
{\texttt{post\_attn\_pre\_norm}} & {Attention block output} & {\-} \\
{\texttt{post\_norm\_pre\_mlp}} & {MLP block input} &{Side Effect Free Steering - App \#6} \\
{\texttt{mlp\_post}} & {MLP block output} & {Hallucination Probe - App \#4} \\
{\texttt{pre\_o\_proj}} & {Attention heads values, before output projection} & {Truthful Intervention - App \#7} \\
{\texttt{post\_qkv}} & {Q, K, V values, before Rotary position embedding} & {\-} \\
{\texttt{pre\_attn}} & {After Rotary position embedding, before attention operation} & {\-} \\
{\texttt{attention scores} (\ac{FA} variant)} & {Inside Attention block, after softmax calculation} & {Universal prefixes - App \#8} \\
{\texttt{router logits}} & {Inside MoE block, before Expert FFN} & {SteerMoE - App \#9} \\
\bottomrule
\end{tabular}
\caption{Submodules exposed by \sys.}
\label{tab:loci}
\end{table*}

\section{All performance metrics}
\label{appx:eval-metrics}

\begin{figure*}[tbp]
  \centering
  \begin{subfigure}[t]{\textwidth}
    \centering
    \includegraphics[height=0.20\textheight]
      {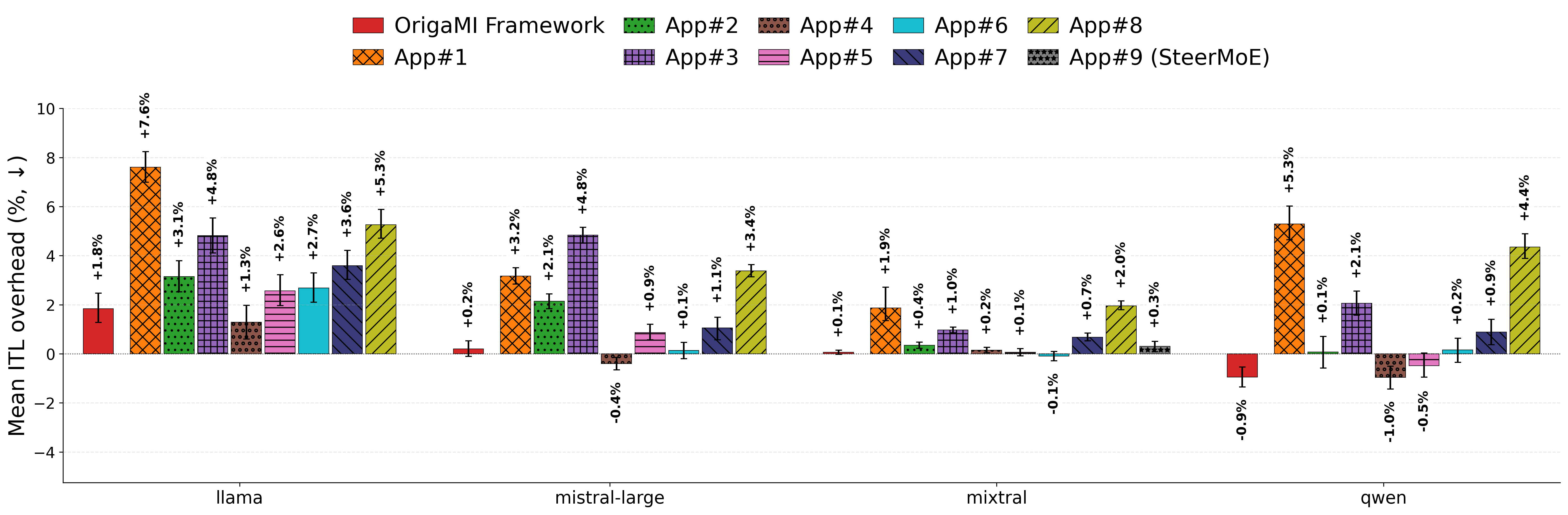}
    \caption{Mean \acs{ITL}}
    \label{fig:eval-combined-mean-itl-a}
  \end{subfigure}
  \vspace{0.5em}
  \begin{subfigure}[t]{\textwidth}
    \centering
    \includegraphics[height=0.20\textheight]
      {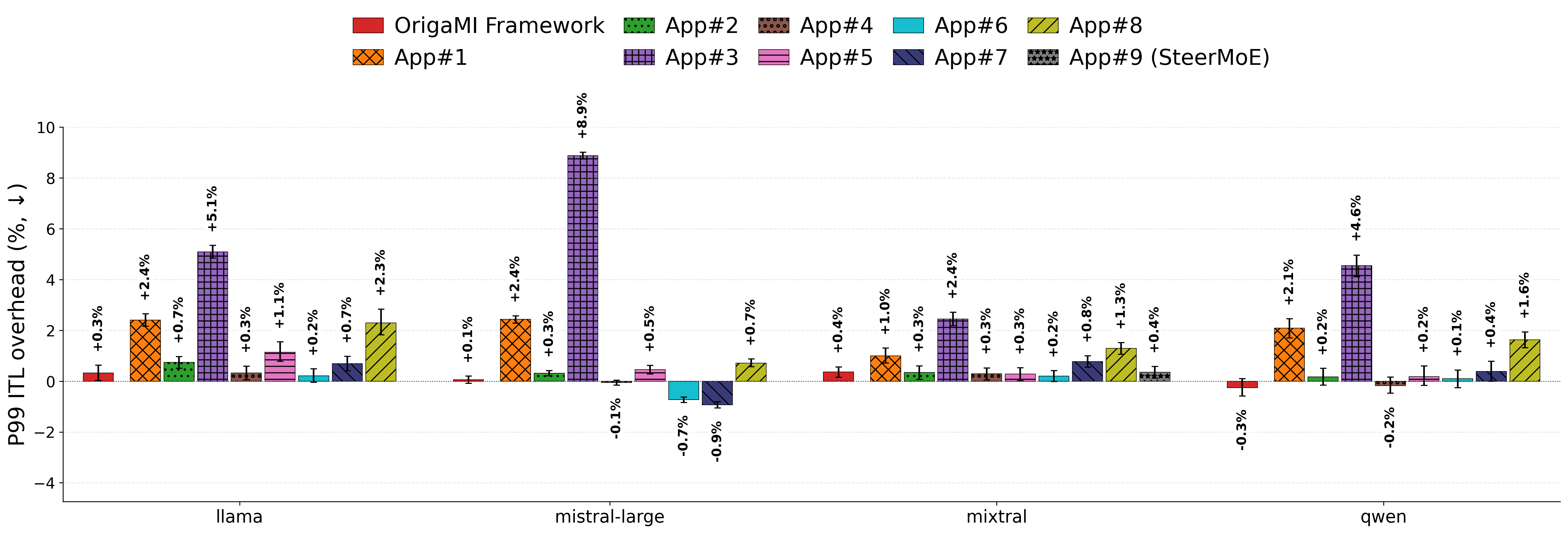}
    \caption{Mean p99 \acs{ITL}}
    \label{fig:eval-combined-p99-itl-b}
  \end{subfigure}
  \vspace{0.5em}
  \begin{subfigure}[t]{\textwidth}
    \centering
    \includegraphics[height=0.20\textheight]
      {Figures/EvalOutput/charts/png_anon/combined_total_token_throughput.png}
    \caption{Mean \acs{TTT}}
    \label{fig:eval-combined-throughput-c}
  \end{subfigure}
  \vspace{0.5em}
  \begin{subfigure}[t]{\textwidth}
    \centering
    \includegraphics[height=0.20\textheight]
      {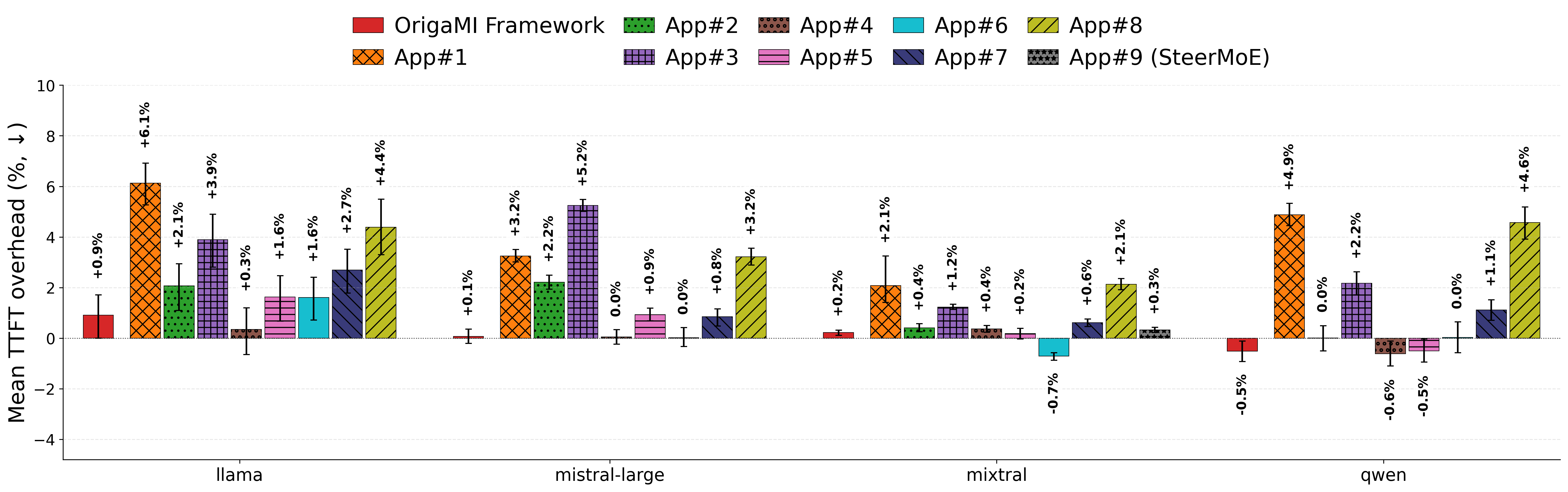}
    \caption{Mean \acs{TTFT}}
    \label{fig:eval-combined-mean-ttft-d}
  \end{subfigure}
  \caption{Overheads relative to the \VLLM baseline for
  App\#1-\#9 deployed with \sys. Each panel reports a per-model breakdown for Llama, Mistral, Mixtral, and
  Qwen3.
  Error bars denote confidence intervals.}
  \label{fig:eval-combined-apps}
\end{figure*}

\cref{fig:eval-combined-apps} extends \cref{fig:eval:apps-ttt} with all
the metrics we measure for \scode{Apps \#1-\#7} on Llama-3.1-8B-Instruct, Mistral-Large-Instruct-123B,
Mixtral-8x7B-Instruct-v0.1, Qwen3-8B: mean \ITL, P99 \ITL, \TTT, and
mean \TTFT, each relative to the \VLLM baseline. The setup and methodology
are those of \cref{SEC:eval:setup}.

\section{Primitive overheads across frameworks}
\label{appx:primitives}

\begin{figure*}[tbp]
  \centering
  \includegraphics[width=\textwidth,height=0.78\textheight,keepaspectratio]
    {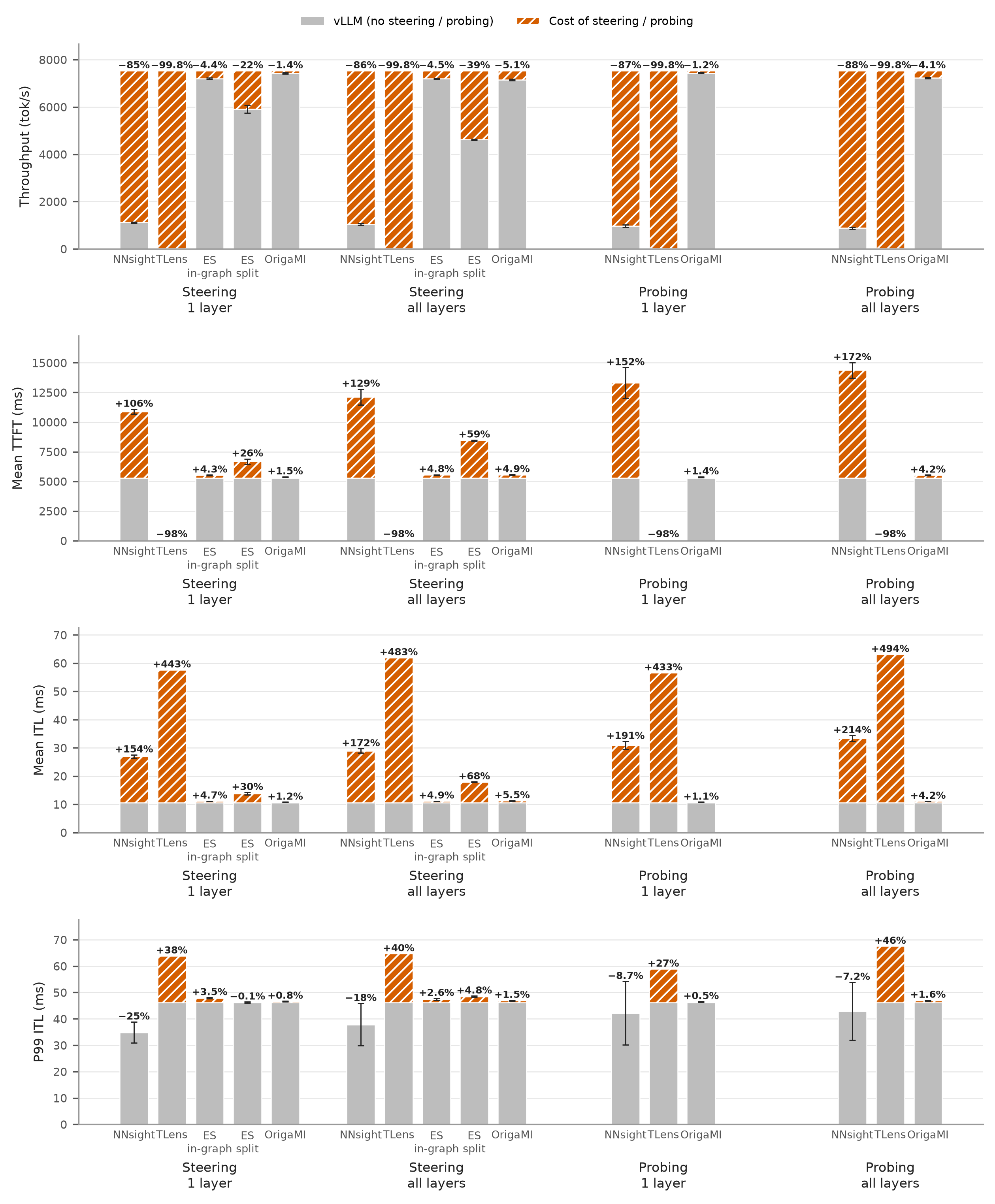}
  \caption{\acs{TTT}, mean \acs{TTFT}, mean \acs{ITL} and P99 \acs{ITL}
  (top to bottom) of steering and probing a single layer and all layers
  of Llama-3.1-8B-Instruct on \NNSIGHT, \TLENS, \ES (ES in-graph, ES
  split) and \sys. Gray: \VLLM without steering or probing; hatched: the
  overhead each system adds, labeled with its change relative to the
  system's own \VLLM baseline. Error bars: 95\% confidence intervals
  (\TLENS~was run once).}
  %\TODO{Check what went wrong with nnsight P99}
  \label{fig:appx:primitives}
\end{figure*}

\cref{fig:appx:primitives} extends \cref{fig:eval:frameworks} and
\cref{fig:eval:easysteer-graphs} with all the metrics we measure for the
steering and probing primitives on Llama-3.1-8B-Instruct: \acs{TTT},
mean \acs{TTFT}, mean \acs{ITL} and P99 \acs{ITL}. We compare \NNSIGHT,
\TLENS, \ES (steering inside the CUDA graph and with a split graph) and
\sys, steering or probing a single layer and all layers. \ES supports
only steering, so it has no probing bars. Each system is measured against
its own \VLLM baseline without steering or probing; the setup and
methodology are those of \cref{SEC:eval:setup}.

\sys adds 0.5\%-5.6\% across all four metrics. \ES with CUDA graphs is
close to it, while splitting the graph raises its mean \acs{TTFT} and
mean \acs{ITL} by up to 59\% and 68\%. \NNSIGHT and \TLENS reduce
throughput by 85\%-99.8\% and increase mean \acs{ITL} by 154\%-494\%.
\TLENS~reports a \acs{TTFT} 98\% lower than \VLLM, so its \acs{TTFT} is
not directly comparable to that of the other systems.
%\TODO{Explain why the TLens TTFT is lower than vLLM. Go through it again} 
The \NNSIGHT~P99
\acs{ITL} varies widely between runs, as its wide confidence intervals
show.

\section{\MIapps on \ES}
\label{appx:easysteer-apps}

\begin{figure}[tbp]
  \centering
  \includegraphics[width=\columnwidth]
    {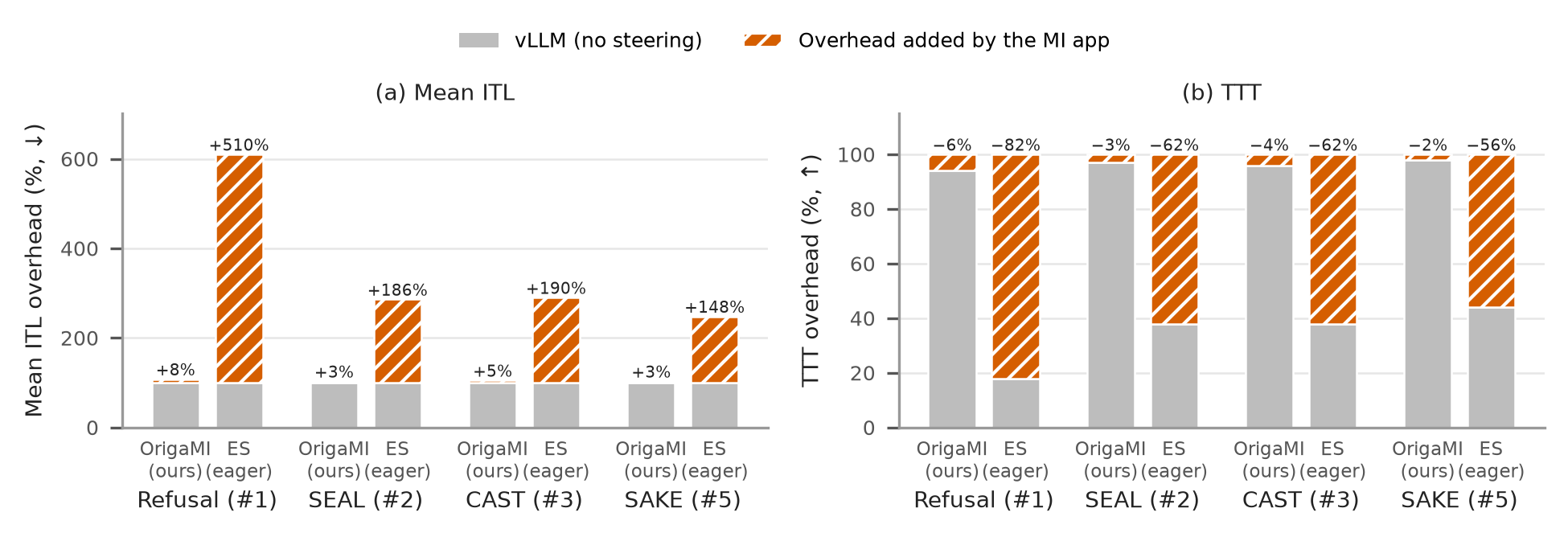}
  \caption{(a) Mean \acs{ITL} and (b) \acs{TTT} of four \MIapps (Refusal,
  SEAL, CAST, SAKE) on Llama-3.1-8B-Instruct, running on \sys and on \ES
  (eager mode), relative to baseline vanilla \VLLM, run with CUDA graph support. Gray: \VLLM without steering; hatched:
  the overhead each system adds. For ITL Lower is better, for TTT higher is better.
  }
  \label{fig:appx:easysteer-apps}
\end{figure}

\cref{fig:appx:easysteer-apps} compares four \MIapps (Refusal - App\#1, SEAL - App\#2, CAST - App\#3,
SAKE - App\#5) running on \sys and on \ES in eager mode, on Llama-3.1-8B-Instruct.
Each system is measured against its own \VLLM baseline without steering; the
setup and methodology are those of \cref{SEC:eval:setup}.
On \ES, the applications increase mean \ac{ITL} by 148--510\% and reduce
\ac{TTT} by 56--82\%, whereas \sys adds at most 8\% to \ac{ITL} and 6\% to
\ac{TTT}.

\ignore{

\section{Per-request application selection}
\label{appx:request-selection}

We tested the scenario where \sys applies an \MIapp only to a subset of the requests in the same batch. We validated 
the correctness of the resulting outputs and verified that  for requests that did not select
an \MIapp, their outputs matched the outputs of the vanilla version. 
}
%\newpage
%\input{Sections/B_EasySteer_Cuda_graph_appendix.tex}

\end{document}